\documentclass[lettersize,journal]{IEEEtran}
\usepackage{amsmath,amsfonts}
\usepackage{array}

\usepackage{textcomp}
\usepackage{stfloats}
\usepackage{verbatim}
\usepackage{cite}
\usepackage{bm}
\usepackage{times}
\usepackage{epsfig}
\usepackage{subfigure}
\usepackage{graphicx}
\usepackage{amsthm}
\usepackage{calc}
\usepackage[ruled,linesnumbered]{algorithm2e}
\usepackage{bbm}
\usepackage{color}
\usepackage{xcolor}
\usepackage{url}
\usepackage{enumitem}
\usepackage{float}
\usepackage{multirow}
\usepackage{multicol}
\usepackage{mathrsfs}
\newtheorem{Definition}{Definition}
\newtheorem{Lemma}{Lemma}

\newcommand{\ie}{\textit{i}.\textit{e}.}

\graphicspath{{./image/}}
\begin{document}

\title{Quaternion Sparse Decomposition for Multi-focus Color Image Fusion}
\author{}
\author{Weihua Yang and Yicong Zhou,~\IEEEmembership{Senior Member,~IEEE}
\thanks{This work was funded by the Science and Technology Development Fund, Macau SAR (File no. 0049/2022/A1, 0050/2024/AGJ), by the University of Macau (File no. MYRG-GRG2024-00181-FST). (Corresponding author: Yicong Zhou.)}
\thanks{Weihua Yang and Yicong Zhou are with the Department of Computer and
Information Science, University of Macau, Macau 999078, China. (e-mail:
weihuayang.um@gmail.com; yicongzhou@um.edu.mo)}}


\markboth{IEEE Transactions on Neural Networks and Learning Systems}%
{Shell \MakeLowercase{\textit{et al.}}: Quaternion Multi-focus Color Image Fusion}

\IEEEpubid{0000--0000/00\$00.00~\copyright~2025 IEEE}

\maketitle

\begin{abstract}
Multi-focus color image fusion refers to integrating multiple partially focused color images to create a single all-in-focus color image. However, existing methods struggle with complex real-world scenarios due to limitations in handling color information and intricate focus information. To address these challenges, this paper proposes a quaternion multi-focus color image fusion framework to perform high-quality color image fusion completely in the quaternion domain. This framework introduces 1) a quaternion consistency-aware focus detection method to jointly learn fine-scale details and structure information of color images and generate patch-wise dual-scale focus maps for high-precision focus detection, 2) a quaternion base-detail fusion strategy to obtain dual-scale initial fusion results across input color images,  and 3) a quaternion structural similarity refinement strategy to adaptively select optimal patches from initial fusion results and produce the final fused result that preserves fine details and spatial consistency. Extensive experiments demonstrate that the proposed framework outperforms state-of-the-art methods.
\end{abstract}


\section{Introduction}
\IEEEPARstart{A}{} single color image typically fails to maintain all scene objects simultaneously in focus due to inherent optical limitations of camera lenses \cite{liu2020multi}. Specifically, when capturing scenes containing objects at varying depths, only regions within the camera's focal plane appear sharp while areas outside this plane inevitably suffer from defocus blur. To overcome this limitation, multi-focus color image fusion (MCIF) techniques integrate multiple partially-focused color images of the same scene by employing focus detection and appropriate fusion strategies to produce a single all-in-focus color image \cite{liu2020multi,liu2023lightweight}. This technology has numerous practical applications, such as digital photography enhancement, preprocessing for subsequent image analysis tasks \cite{li2017pixel}, image segmentation \cite{liu2023lightweight} and object recognition \cite{li2024samf}.
 
Recent advances in deep learning have significantly promoted MCIF development to enhance multi-scale feature extraction capabilities and improve focused-region discrimination through decision map-based networks \cite{zhao2021depth, wang2021mfif} and end-to-end architectures \cite{zhang2024exploit, xie2024swinmff,li2024focus}. However, it is impractical to acquire ideal ground truth (all-in-focus color images) in real-world conditions \cite{zhao2024image}. These deep-learning-based methods typically rely on synthetic training datasets for training \cite{wang2022self}. Consequently, their generalization capability on real-world images remains limited.

Conventional MCIF methods depend only on human prior knowledge to identify focused objects without the training phase \cite{wang2023multi}. 
These methods commonly extend grayscale fusion strategies directly to color images either using grayscale conversion \cite{qiu2019guided} or channel-wise processing \cite{yang2009multifocus}. Grayscale-conversion-based MCIF methods first convert the input color images into grayscale images and perform intensity-driven focus detection to obtain focus maps from the grayscale inputs for guiding the fusion process of original color images \cite{qiu2019guided}. For instance, \cite{chen2021multi} averaged color channels into a grayscale image and obtained focus maps by multi-scale morphological filtering. \cite{roy2020scheme} transformed the color image into the YUV color space and detected focus levels employing the zero crossing and canny edge detectors. However, these methods may overlook multi-channel sharpness variations and cross-channel correlations. This leads to inaccurate focus detection, ghosting artifacts, and blurred boundaries in fused color results \cite{xiao2019multi,li2023generation}.

Channel-wise processing-based MCIF methods treat RGB channels independently and apply grayscale fusion methods separately to each channel.
For instance, Yang et al. \cite{yang2009multifocus} applied sparse coding techniques to MCIF and separately measured focus levels in each channel and fused sparse coefficients channel-by-channel. They fail to consider the cross-channel correlations and introduce spectral-spatial inconsistencies and undesirable color distortions in the fused results \cite{yu2013quaternion}. 

\IEEEpubidadjcol Quaternion representation as a promising tool of color image processing can overcome the aforementioned limitations of color image processing in conventional MCIF methods \cite{liu2014region,miao2023quaternion}. It effectively captures color information and preserves cross-channel correlations by treating RGB channels as imaginary components in the quaternion domain \cite{li2022automatic, xiao2018two}. Recently, a quaternion higher-order singular value decomposition (QHOSVD) model \cite{miao2023quaternion} was developed by modeling multi-focus color images as third-order quaternion tensors and successfully maintained cross-channel consistency in the stages of focus detection and fusion. However, QHOSVD selects fusion patches based on singular-value energy. This may result in spatial inconsistency and degrade the visual quality of the fused color images.

Motivated by these critical issues, we propose a quaternion multi-focus color image fusion (QMCIF) framework to explicitly designed to simultaneously ensure high-quality multi-focus color image fusion in both spatial and color channels dimensions.
Our main contributions are presented as follows:
\begin{itemize}
    \item We propose a QMCIF framework to perform color image fusion completely in the quaternion domain. QMCIF is able to obtain high-quality fused color images under various complex scenarios.
    \item To effectively measure the focus level and enhance the capability of focus detection in uncertain regions, QMCIF introduces a quaternion consistency-aware focus detection method to explicitly learn focus features of color images and derive dual-scale focus maps in both high-texture and low-gradient regions.
    \item To preserve color structure and detail information, QMCIF introduces a patch-wise quaternion base-detail fusion strategy to fuse partially focused color images in detail-scale and base-scale individually.  
    \item To balance the trade-off between focus information preservation and artifact removal, QMCIF further introduces a quaternion structural similarity refinement strategy to produce the final high-quality fused color image. 
    \item Extensive experiments on various challenging scenarios demonstrate that QMCIF outperforms the state-of-the-art methods.
\end{itemize}
The rest of this paper is organized as follows: Section \ref{overview} presents the preliminaries of quaternion algebra. Section \ref{framework:QMCIF} introduces our framework. Section \ref{experiments} presents the experiments and comparisons. Finally, Section \ref{conclusion} gives the conclusions.
\section{Preliminaries} \label{overview}
This section presents quaternion representations of scalar numbers, vectors and matrices in detail. Table \ref{tab:table1} shows the main notations and mathematic symbols used in this paper. 
\begin{table}[htbp]
\caption{Notations and mathematical symbols. }
    \centering \scalebox{0.8}[0.8]{
    \begin{tabular}{|c|l|}
    \hline
         $\mathbbm{R}$, $\mathbbm{H}$& real space, quaternion space\\
         \hline
         $a$, $\boldsymbol{\mathrm{a}}$, $\boldsymbol{\mathrm{A}}$ & real scalar number, vector, matrix\\
         \hline
         $\dot{a}$, $\dot{\boldsymbol{\mathrm{a}}}$, $\dot{\boldsymbol{\mathrm{A}}}$ & quaternion scalar number, vector, matrix \\
         \hline
         ${\boldsymbol{\mathrm{I}}}_{d}$, $\dot{\boldsymbol{\mathrm{I}}}_{d}$ & real identity matrix, quaternion identity matrix\\
         \hline
         ${(\cdot)}^{\mathrm{T}}$,$\overline{(\cdot)}$,${(\cdot)}^{\mathrm{H}}$, ${(\cdot)}^{-1}$  & transpose, conjugate , conjugate transpose and inverse representation\\
         \hline
         ${\Vert \cdot \Vert}_1$, ${\Vert \cdot \Vert}_F$, ${\Vert \cdot \Vert}_*$ & ${\ell}_1$ norm, Frobenius norm, nuclear norm\\
         \hline
    \end{tabular} 
    }
    \label{tab:table1}
\end{table}

The set of quaternions $\mathbbm{H}$ defines a 4-components normed algebra \cite{flamant2021general} over the real numbers $\mathbbm{R}$ (\ie,  basis $\{1,i,j,k\}$) as follows:
\begin{equation*}
    \begin{split}
     \dot{q}=q_a+q_b i+q_c j+ q_d  k,
    \end{split}
    \label{eq1}
\end{equation*}
where $q_a,q_b,q_c,q_d \in \mathbbm{R}$ are components of $\dot{q}$. $i,j,k$ are the imaginary parts such that
\begin{equation*}
    \begin{split}
i^2=j^2=k^2=ijk=-1,ij=-ji,ij=k
    \end{split}
    \label{eq2}
\end{equation*}
These relations imply that quaternion multiplication is non-commutative. For $\dot{q},\dot{p} \in \mathbbm{H}$, $\dot{q}\dot{p}\neq \dot{p}\dot{q}$.

A quaternion $\dot{q}$ is a pure quaternion number if its real part $q_a=0$, namely $\dot{q}=q_bi+q_cj+q_dk$.
Next, we will present several definitions that will be used in this paper.
\begin{Definition}
Given a quaternion vector $\dot{\boldsymbol{\mathrm{q}}}=(\dot{q}_s)\in \mathbbm{H}^M$, and a quaternion matrix $\dot{\boldsymbol{\mathrm{A}}}=({\dot{a}}_{s,t})\in \mathbbm{H}^{M\times N}$, where $s=1,\cdots, M$ and $t=1,\cdots, N$ are the row and column indices respectively.
\begin{enumerate}[itemindent=1.5em,label=\emph{\arabic*)}]
    \item conjugate transpose: $\dot{{\boldsymbol{\mathrm{A}}}}^{\mathrm{H}}=(\overline{\dot{a}_{t,s}}) \in \mathbbm{H}^{N\times M}$
    \item ${\ell}_1$ norm: ${\Vert \dot{\boldsymbol{\mathrm{q}}} \Vert}_1=\sum_{s=1}^{M}{\vert \dot{q}_s\vert}$
    \item ${\ell}_2$ norm: ${\Vert \dot{\boldsymbol{\mathrm{q}}} \Vert}_2={(\sum_{s=1}^{M}{\vert \dot{q}_s\vert}^2)}^{\frac{1}{2}}$
    \item $Frobenius$  norm: ${\Vert \dot{\boldsymbol{\mathrm{A}}} \Vert}_F={(\sum_{s=1}^{M}\sum_{t=1}^{N}{{\vert \dot{a}_{s,t}\vert}^2})}^{\frac{1}{2}}$
\end{enumerate}
\end{Definition}
\begin{Definition}
The rank of a quaternion matrix $\dot{\boldsymbol{\mathrm{A}}}$ is r if and only if $\dot{\boldsymbol{\mathrm{A}}}$ has r nonzero singular values {\rm \cite{chen2019low}}.
\end{Definition}
\begin{Definition}
Given a color image $\boldsymbol{\mathrm{I}}$ in three-dimensional real space, its quaternion representation is defined as:
\begin{equation}
    \begin{split}
     \dot{\boldsymbol{\mathrm{I}}}(s,t)
     =\boldsymbol{\mathrm{I}}_r(s,t) i+\boldsymbol{\mathrm{I}}_g(s,t) j+ \boldsymbol{\mathrm{I}}_b(s,t) k
    \end{split}
    \label{eq3}
\end{equation}
where $\dot{\boldsymbol{\mathrm{I}}}(s,t)$ is the quaternion representation of the color image pixel at the location of $(s,t)$. $\boldsymbol{\mathrm{I}}_r$, $\boldsymbol{\mathrm{I}}_g$ and $\boldsymbol{\mathrm{I}}_b$ is red, green and blue channels in $\boldsymbol{\mathrm{I}}$ respectively.
\end{Definition}
\begin{Definition}
(Quaternion derivatives \cite{xu2015optimization}) The quaternion derivatives of the real scalar function $f$ : $\mathbbm{H}^{M \times N} \rightarrow \mathbbm{R}$ with respect to $\dot{\boldsymbol{\mathrm{Q}}}$ $\in \mathbbm{H}^{M \times N}$ are defined by
\begin{equation*}
    \frac{\partial f}{\partial \dot{\boldsymbol{\mathrm{Q}}}}=\begin{pmatrix}
        \frac{\partial f}{\partial {\dot{q}}_{1,1}} & \cdots & \frac{\partial f}{\partial {\dot{q}}_{1,N}}\\
        \vdots & \ddots &\vdots \\
        \frac{\partial f}{\partial {\dot{q}}_{M,1}} & \cdots & \frac{\partial f}{\partial {\dot{q}}_{M,N}}.
    \end{pmatrix}
\end{equation*}\label{dfn4}
where $\frac{\partial f}{\partial {\dot{q}}}=\frac{1}{4}(\frac{\partial f}{\partial {q_a}}-\frac{\partial f}{\partial {q_b}}i-\frac{\partial f}{\partial {q_c}}j-\frac{\partial f}{\partial {q_d}}k)$.
The quaternion derivation has the following properties. They make the quaternion derivation significantly different from real-valued and complex-valued ones.
\begin{enumerate}
    \item Non-commutativity: $\frac{\partial f}{\partial q_b}$, $\frac{\partial f}{\partial q_c}$, $\frac{\partial f}{\partial q_d}$ cannot be swapped with $i$, $j$, $k$.
    \item $\frac{\partial {\dot{q}}}{\partial {\dot{q}}}=\frac{\partial \overline{\dot{q}}}{\partial \overline{\dot{q}}}=1$, $\frac{\partial {\dot{q}}}{\partial \overline{\dot{q}}}=\frac{\partial \overline{\dot{q}}}{\partial {\dot{q}}}=-\frac{1}{2}$.
    \item Product rule: $\frac{\partial fg}{\partial {\dot{q}}}=f\frac{\partial g}{\partial {\dot{q}}}+\frac{\partial f}{\partial {\dot{q}}^{f}}g\neq f\frac{\partial g}{\partial {\dot{q}}}+\frac{\partial f}{\partial {\dot{q}}}g$.
    \item Chain rule: $\frac{\partial f(g)}{\partial {\dot{q}}}=\frac{\partial f}{\partial g}\frac{\partial g}{\partial {\dot{q}}}+\frac{\partial f}{\partial g^{\boldsymbol{i}}}\frac{\partial g^{\boldsymbol{i}}}{\partial {\dot{q}}}+\frac{\partial f}{\partial g^{\boldsymbol{j}}}\frac{\partial g^{\boldsymbol{j}}}{\partial {\dot{q}}}+\frac{\partial f}{\partial g^{\boldsymbol{k}}}\frac{\partial g^{\boldsymbol{k}}}{\partial {\dot{q}}}$.
\end{enumerate}
\end{Definition}
\begin{Definition}
    Quaternion Structural Similarity ($Q_{SSIM}$) \cite{kolaman2011quaternion} measures the similarity between two color images in the quaternion domain. It is defined by
\begin{equation}
    Q_{SSIM}({\dot{\boldsymbol{\mathrm{X}}}},{\dot{\boldsymbol{\mathrm{Y}}}})=\overline{(\dot{a})}\dot{b},\label{eq-QSSIM}
\end{equation}
where 
\begin{equation*}
    \begin{aligned}
        \dot{a}=\frac{2\overline{\dot{\mu}_{\dot{\boldsymbol{\mathrm{X}}}}}\dot{\mu}_{\dot{\boldsymbol{\mathrm{Y}}}}+C_1}{\overline{{\dot{\boldsymbol{\mathrm{\mu}}}}_{\dot{\boldsymbol{\mathrm{X}}}}}{\dot{\mu}}_{\dot{\boldsymbol{\mathrm{X}}}}+\overline{{\dot{\mu}}_{\dot{\boldsymbol{\mathrm{Y}}}}}{\dot{\mu}}_{\dot{\boldsymbol{\mathrm{Y}}}}+C_1}, & \dot{b}=\frac{2\overline{\dot{\sigma}_{\dot{\boldsymbol{\mathrm{X}}}}}\dot{\sigma}_{\dot{\boldsymbol{\mathrm{Y}}}}+C_2}{\overline{{\dot{\boldsymbol{\mathrm{\sigma}}}}_{\dot{\boldsymbol{\mathrm{X}}}}}{\dot{\sigma}}_{\dot{\boldsymbol{\mathrm{X}}}}+\overline{{\dot{\sigma}}_{\dot{\boldsymbol{\mathrm{Y}}}}}{\dot{\sigma}}_{\dot{\boldsymbol{\mathrm{Y}}}}+C_2}.
    \end{aligned}
\end{equation*}
$\dot{\boldsymbol{\mathrm{X}}}$ and $\dot{\boldsymbol{\mathrm{Y}}}$ are input patches. $\dot{\mu}_{\dot{\boldsymbol{\mathrm{X}}}}$ and $\dot{\sigma}_{\dot{\boldsymbol{\mathrm{X}}}}$ are denoted as mean and variance values of the image patch $\dot{\boldsymbol{\mathrm{X}}}$. Constants $C_1$ and $C_2$ are set as small as possible.
\label{dfn5}
\end{Definition}
\begin{figure*}[htbp]
\centering
\includegraphics[width=1\textwidth]{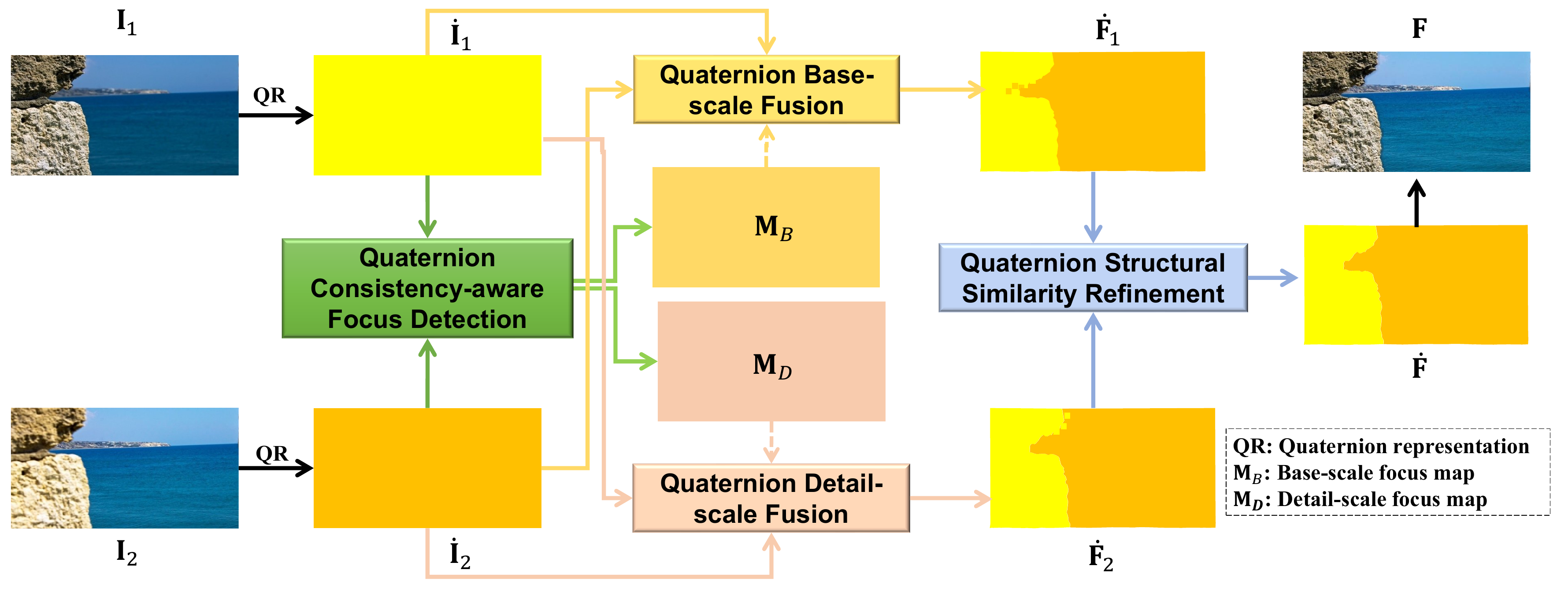}
\caption{Flowchart of quaternion multi-focus color image fusion (QMCIF) framework.} 
\label{fig:Framework}
\end{figure*}
\begin{Lemma}
    (Quaternion nuclear norm \cite{chen2019low}) For any $\lambda \geq 0$, quaternion matrix $\dot{\boldsymbol{\mathrm{Y}}}$ and $\dot{\boldsymbol{\mathrm{X}}}$ $\in \mathbbm{H}^{M\times N}$ both with the rank of $r$, the quaternion nuclear norm can be defined as
    \begin{equation*}
\mathop{\arg\min}_{\dot{\boldsymbol{\mathrm{X}}}}\ \  \frac{1}{2}{\Vert \dot{\boldsymbol{\mathrm{Y}}}-\dot{\boldsymbol{\mathrm{X}}} \Vert}_{F}^2+ \lambda{\Vert \dot{\boldsymbol{\mathrm{X}}} \Vert}_*.
    \end{equation*}
    Its closed form solution is $\dot{\boldsymbol{\mathrm{U}}} S_\lambda(\Sigma)\dot{\boldsymbol{\mathrm{V}}}^H$ where $\dot{\boldsymbol{\mathrm{U}}}$ $\in \mathbbm{H}^{M\times r} $ and $\dot{\boldsymbol{\mathrm{V}}}$ $\in \mathbbm{H}^{r\times N} $ are the orthogonal quaternion matrices, $\Sigma $ is a diagonal real matrix with singular values and $S_\lambda(\Sigma)$ denotes the soft thresholding operator with parameter $\lambda$.\label{lma1}
\end{Lemma}
\begin{Lemma}
    (Quaternion soft-thresholding operator \cite{xiao2018two}) Given $\dot{\boldsymbol{\mathrm{X}}}$ and $ \dot{\boldsymbol{\mathrm{Y}}}\in \mathbbm{H}^{M\times N}$, we solve the quaternion ${\ell}_1$ norm optimization problem with respect to $\dot{\boldsymbol{\mathrm{X}}}$  using a quaternion soft-thresholding operator.
    \begin{align*}        &\dot{\boldsymbol{\mathrm{X}}}=\mathop{\arg\min}_{\dot{\boldsymbol{\mathrm{X}}}}\ \ \tau{\| \dot{\boldsymbol{\mathrm{X}}} \|}_{1}
        +\frac{1}{2}\|\dot{\boldsymbol{\mathrm{X}}}-\dot{\boldsymbol{\mathrm{Y}}} \|{_F^2}, \\
        &\dot{\boldsymbol{\mathrm{X}}}(:,i)=\left \{ 
    \begin{aligned}
        &\frac{{\| \dot{\boldsymbol{\mathrm{Y}}}(:,i)\|}_1-\tau}{{\| \dot{\boldsymbol{\mathrm{Y}}}(:,i)\|}_1}\dot{\boldsymbol{\mathrm{Y}}}(:,i),& &{\| \dot{\boldsymbol{\mathrm{Y}}}(:,i)\|}_1>\tau\\
        &0,& & {\rm otherwise}
    \end{aligned}
    \right.
    \end{align*}
    \label{lma2}
\end{Lemma}
\section{Proposed Framework} \label{framework:QMCIF}
This section presents our quaternion multi-focus color image fusion (QMCIF) framework in detail.  Section A provides an overview of our QMCIF framework. Section B introduces our quaternion consistency-aware focus detection method. Section C presents our quaternion base-detail fusion strategy. Section D presents our quaternion structural similarity refinement strategy.
\subsection{Overview}\label{flowchart}
Fig.~\ref{fig:Framework} illustrates our quaternion multi-focus color image fusion (QMCIF) framework. Unlike most existing methods that process color images in a grayscale manner, QMCIF performs joint fusion across all color channels within a unified quaternion representation. Input color images $\boldsymbol{\mathrm{I}}_1$, $\boldsymbol{\mathrm{I}}_2$ are converted into their quaternion representations $\dot{\boldsymbol{\mathrm{I}}}_1$, $\dot{\boldsymbol{\mathrm{I}}}_2$ using Eq. \eqref{eq3}. A quaternion consistency-aware focus detection method jointly estimates their optimal coefficient matrices and detail layers to construct dual-scale focus maps for accurate focus estimation. Under the guidance of dual-scale focus maps, a patch-wise quaternion fusion strategy is applied to fuse the quaternion representations $\dot{\boldsymbol{\mathrm{I}}}_1$ and $\dot{\boldsymbol{\mathrm{I}}}_2$ at the base and detail scales to generate the initial fusion results $\dot{\boldsymbol{\mathrm{F}}}_1$ and $\dot{\boldsymbol{\mathrm{F}}}_2$. They are further applied with a quaternion structure-similarity-based refinement strategy patch-by-patch to generate the final fused result $\dot{\boldsymbol{\mathrm{F}}}$ that is converted back into the real domain to obtain a high-quality all-in-focus color image. Notably, in additional to fuse two color images at a time, our QMCIF framework can be applied to fuse multiple input color images simultaneously. This advantage is verified by the experiments in Figs. \ref{lytro3} and \ref{mffw14}.

\subsection{Quaternion consistency-aware focus detection}\label{methods}
\begin{figure}[htbp]
\centering {\includegraphics[width=1\columnwidth]{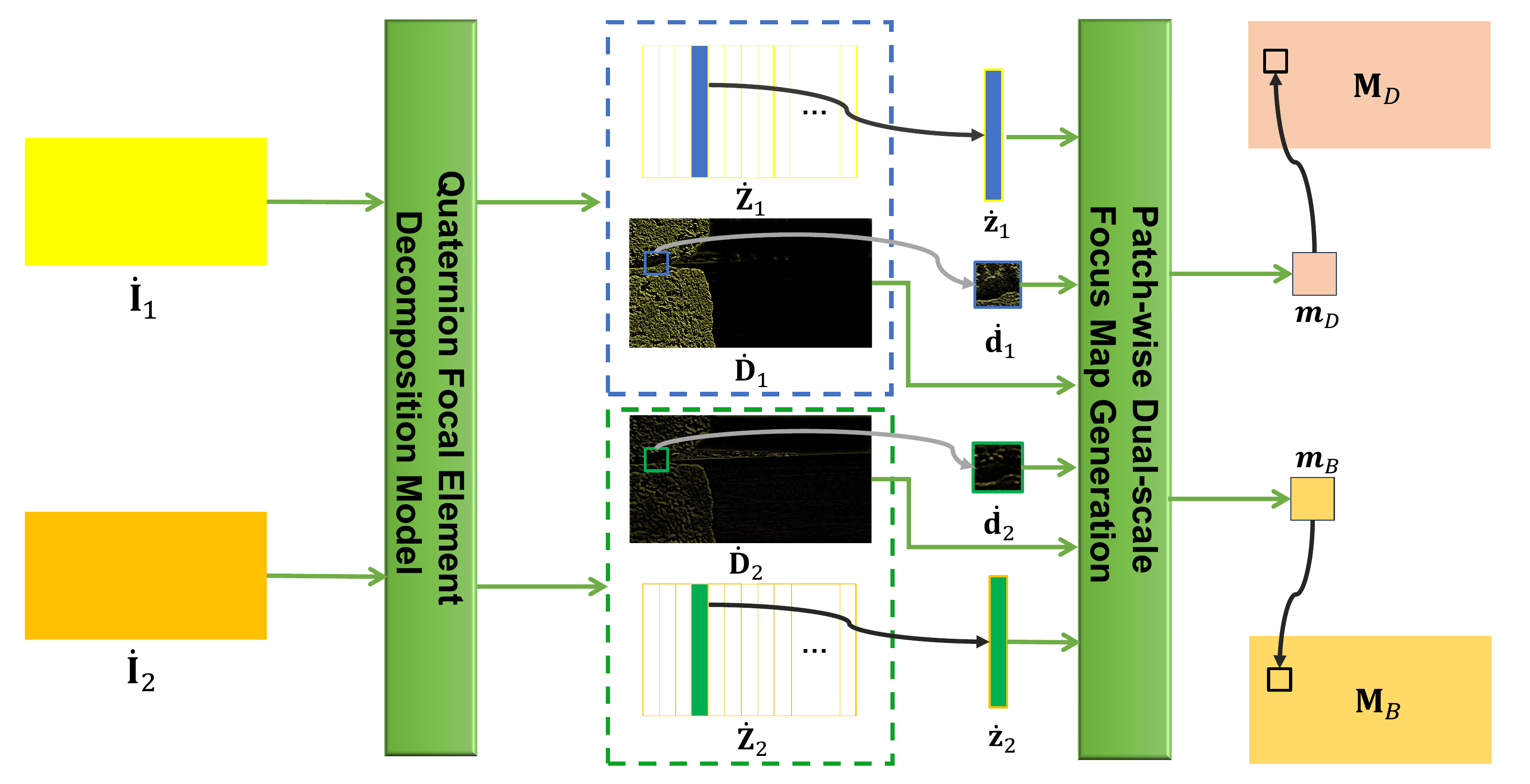}}
\caption{Our quaternion consistency-aware focus detection (QCAFD).} \label{fig:QCAFD-flowchart}
\end{figure}

To achieve high-precision focus estimation in both high-texture and low-gradient regions, our quaternion consistency-aware focus detection method contains two main steps: (1) a quaternion focal element decomposition model to effectively extract focus-related features in the quaternion domain; and (2) a patch-wise dual-scale focus map generation strategy. Its flowchart is shown in Fig. \ref{fig:QCAFD-flowchart}.
\subsubsection{Quaternion focal element decomposition} \label{model}
Given an input quaternion representation $\dot{\boldsymbol{\mathrm{I}}}$ $\in \mathbbm{H}^{M\times N}$, our quaternion focal element decomposition (QFED) model decomposes texture information from its static background as a strong focus measure. Our QFED can be formulated as follows.
\begin{equation}
    \begin{aligned}
\mathop{\arg\min}_{\dot{\boldsymbol{\mathrm{D}}},\dot{\boldsymbol{\mathrm{Z}}}} \ \ &\sum_{k=1}^{K} {\| \dot{\boldsymbol{\mathrm{Z}}}_k \|}_{*}+\alpha ({\| \nabla_1 \dot{\boldsymbol{\mathrm{B}}} \|}_1+{\| \nabla_2 \dot{\boldsymbol{\mathrm{B}}} \|}_1)\\
    & + \beta \|\dot{\boldsymbol{\mathrm{D}}} \|_1 +\lambda \| \dot{\boldsymbol{\mathrm{E}}}\|_{F}^{2}, \\
    &s.t. \ \ \dot{\boldsymbol{\mathrm{I}}}=\dot{\boldsymbol{\mathrm{B}}}+\dot{\boldsymbol{\mathrm{D}}}+\dot{\boldsymbol{\mathrm{E}}},\ \mathcal{R}(\dot{\boldsymbol{\mathrm{B}}})_{k}=\dot{\boldsymbol{\mathrm{A}}}\dot{\boldsymbol{\mathrm{Z}}}_{k}
    \label{eq6}
    \end{aligned}
\end{equation}
where ${\nabla}_1$ and ${\nabla}_2$ represent the quaternion vertical and horizontal gradients respectively. Operator $\mathcal{R}(\cdot)$ extracts image patches of size $\sqrt{d} \times \sqrt{d}$ and stacks their vectorized forms into a quaternion matrix in $ \mathbbm{H}^{d \times P}$. Column vectors are divided into $K$ groups denoted as $\mathcal{R}(\dot{\boldsymbol{\mathrm{B}}})=\{ \mathcal{R}(\dot{\boldsymbol{\mathrm{B}}})_{1}, \mathcal{R}(\dot{\boldsymbol{\mathrm{B}}})_{2}, \cdots, \mathcal{R}(\dot{\boldsymbol{\mathrm{B}}})_{K} \}$ where $\mathcal{R}(\dot{\boldsymbol{\mathrm{B}}})_{k}$ represents the $k^{th}$ group. $\dot{\boldsymbol{\mathrm{A}}}\dot{\boldsymbol{\mathrm{Z}}}{_k}$ captures the low-rank property of $\mathcal{R}(\dot{\boldsymbol{\mathrm{B}}})_{k}$. ${\| \dot{\boldsymbol{\mathrm{Z}}}_k \|}_{*}$ serves as a convex relaxation of the rank function.
The positive parameters $\alpha$, $\beta$ and $\lambda$ balance the effects of each term in the optimization process. 

Our QFED decomposes $\dot{\boldsymbol{\mathrm{I}}}$ into a base-scale layer $\dot{\boldsymbol{\mathrm{B}}}$ with low-rank structural information, a detail-scale layer $\dot{\boldsymbol{\mathrm{D}}}$ with sparse details and a Gaussian noise layer $\dot{\boldsymbol{\mathrm{E}}}$. To fully utilize the focus information in $\dot{\boldsymbol{\mathrm{B}}}$, QFED uses nonlocal similarity to extract local patches in $\dot{\boldsymbol{\mathrm{B}}}$, stacks these patches into column vectors, and partitions them into several groups using K-means clustering method. QFED learns a set of coefficient matrices $\dot{\boldsymbol{\mathrm{Z}}}_{k}$ from these groups using pretrained quaternion dictionary $\dot{\boldsymbol{\mathrm{A}}}$ $\in \mathbbm{H}^{d \times L}$ and concatenates these coefficient matrices to obtain $\dot{\boldsymbol{\mathrm{Z}}}$ $\in \mathbbm{H}^{L \times P}$. This ensures that patches within the same group are similar. 
 

\textbf{Optimization}.
To solve the optimization problem of our QFED in Eq. \eqref{eq6}, we here introduce an iterative learning algorithm under the framework of the quaternion alternating direction method of multipliers \cite{flamant2021general}. According to the definitions and lemmas in Section \ref{overview}, we write the Lagrangian function of our QFED in Eq. \eqref{eq6} as follows:
\begin{equation}
    \begin{aligned}
    \mathcal{L}&=\sum_{k=1}^{K} {\| \dot{\boldsymbol{\mathrm{J}}}_k \|}_{*}+\left \langle \dot{\boldsymbol{\mathrm{Y}}}{_{1,k}},\ \dot{\boldsymbol{\mathrm{Z}}}{_k}-\dot{\boldsymbol{\mathrm{J}}}{_k} \right \rangle 
    +\frac{\mu}{2}\|\dot{\boldsymbol{\mathrm{Z}}}{_k}-\dot{\boldsymbol{\mathrm{J}}}{_k} \|_F^2\\ 
    &+\left \langle \dot{\boldsymbol{\mathrm{Y}}}{_{2,k}},\ \mathcal{R}(\dot{\boldsymbol{\mathrm{B}}}){_k}-\dot{\boldsymbol{\mathrm{A}}}\dot{\boldsymbol{\mathrm{Z}}}{_k} \right \rangle+\frac{\mu}{2}\|\mathcal{R}(\dot{\boldsymbol{\mathrm{B}}}){_k}-\dot{\boldsymbol{\mathrm{A}}}\dot{\boldsymbol{\mathrm{Z}}}{_k} \|_F^2\\
    &+ \alpha\|\dot{\boldsymbol{\mathrm{G}}}_{1} \|_1
    +\left \langle \dot{\boldsymbol{\mathrm{Y}}}{_3},\ \dot{\boldsymbol{\mathrm{G}}}_{1}-\nabla_1 \dot{\boldsymbol{\mathrm{B}}} \right \rangle
    +\frac{\mu}{2}\|\dot{\boldsymbol{\mathrm{G}}}_{1}-\nabla_1 \dot{\boldsymbol{\mathrm{B}}} \|_F^2  \\
    &+ \alpha\|\dot{\boldsymbol{\mathrm{G}}}_{2} \|_1+\left \langle\dot{\boldsymbol{\mathrm{Y}}}{_4},\ \dot{\boldsymbol{\mathrm{G}}}_{2}-\nabla_2 \dot{\boldsymbol{\mathrm{B}}}\right \rangle
    +\frac{\mu}{2}\|\dot{\boldsymbol{\mathrm{G}}}_{2}-\nabla_2 \dot{\boldsymbol{\mathrm{B}}} \|_F^2\\
    &+ \beta \|\dot{\boldsymbol{\mathrm{D}}} \|_1 +\lambda \| \dot{\boldsymbol{\mathrm{E}}}\|_{F}^2+\left \langle \dot{\boldsymbol{\mathrm{Y}}}{_5},\ \dot{\boldsymbol{\mathrm{I}}}-\dot{\boldsymbol{\mathrm{B}}}-\dot{\boldsymbol{\mathrm{D}}}-\dot{\boldsymbol{\mathrm{E}}} \right \rangle\\
    &+\frac{\mu}{2}\| \dot{\boldsymbol{\mathrm{I}}}-\dot{\boldsymbol{\mathrm{B}}}-\dot{\boldsymbol{\mathrm{D}}}-\dot{\boldsymbol{\mathrm{E}}} \|_F^2.
    \end{aligned}
    \label{eq8}
\end{equation}
where ${\{{\dot{\boldsymbol{\mathrm{J}}}}_k\}}_{k=1}^K \in \mathbbm{H}^{L \times P }$ represent a set of auxiliary variables to replace ${\{{\dot{\boldsymbol{\mathrm{Z}}}}_k\}}_{k=1}^K$. Matrices $ {\dot{\boldsymbol{\mathrm{G}}}}_1$ and ${\dot{\boldsymbol{\mathrm{G}}}}_2 \in \mathbbm{H}^{M \times N}$ correspond to quaternion horizontal and vertical gradient matrices, replacing $\nabla_1 \dot{\boldsymbol{\mathrm{B}}}$ and $\nabla_2 \dot{\boldsymbol{\mathrm{B}}}$ respectively. Variables ${ \{{\dot{\boldsymbol{\mathrm{Y}}}}_{1,k}\}}_{k=1}^{K}$ $\in \mathbbm{H}^{L \times P}$, ${\{{\dot{\boldsymbol{\mathrm{Y}}}}_{2,k} \}}_{k=1}^{K}$ $\in \mathbbm{H}^{d \times P}$, ${\dot{\boldsymbol{\mathrm{Y}}}}_{3}, {\dot{\boldsymbol{\mathrm{Y}}}}_{4}$ and ${\dot{\boldsymbol{\mathrm{Y}}}}_{5}\in \mathbbm{H}^{M \times N}$ are the quaternion Lagrangian multipliers in the optimization process. $\left \langle \cdot \right \rangle$ denotes a quaternion trace product while $\mu$ is a penalty factor.
Variables $\dot{\boldsymbol{\mathrm{J}}}{_k}$, $\dot{\boldsymbol{\mathrm{Z}}}{_k}$, $\dot{\boldsymbol{\mathrm{Y}}}{_{1,k}}$ and $\dot{\boldsymbol{\mathrm{Y}}}{_{2,k}}$, $\dot{\boldsymbol{\mathrm{G}}}{_1}$, $\dot{\boldsymbol{\mathrm{G}}}{_2}$, $\dot{\boldsymbol{\mathrm{D}}}$, $\dot{\boldsymbol{\mathrm{E}}}$, $\dot{\boldsymbol{\mathrm{Y}}}{_3}$, $\dot{\boldsymbol{\mathrm{Y}}}{_4}$, $\dot{\boldsymbol{\mathrm{Y}}}{_5}$ are updated in each iteration.

\textbf{Update ${\{\dot{\boldsymbol{\mathrm{J}}}{_k}\}}_{k=1}^{K}$}. Fix ${\{\dot{\boldsymbol{\mathrm{Z}}}{_k}\}}_{k=1}^{K}$ and ${\{\dot{\boldsymbol{\mathrm{Y}}}{_{1,k}}\}}_{k=1}^{K}$. Let $\dot{\boldsymbol{\mathrm{P}}}=$ $\dot{\boldsymbol{\mathrm{Z}}}{_k}+\frac{\dot{\boldsymbol{\mathrm{Y}}}{_{1,k}}}{\mu}$. The subproblem of $\dot{\boldsymbol{\mathrm{J}}}{_k}$ is reduced to:
\begin{equation}
    \begin{aligned}
        \dot{\boldsymbol{\mathrm{J}}}{_k}
    =\mathop{\arg\min}_{\dot{\boldsymbol{\mathrm{J}}}{_k}}\ \ \frac{1}{\mu}{\| \dot{\boldsymbol{\mathrm{J}}}_k \|}_{*}
    +\frac{1}{2}\|\dot{\boldsymbol{\mathrm{J}}}{_k}-\dot{\boldsymbol{\mathrm{P}}} \|_F^2. 
    \end{aligned}
    \label{eq20}
\end{equation}
Eq. \eqref{eq20} is solved using the Lemma \ref{lma1} in Section~\ref{overview}.

\textbf{Update ${\{\dot{\boldsymbol{\mathrm{Z}}}{_k}\}}_{k=1}^{K}$.}  Fix ${\{\dot{\boldsymbol{\mathrm{B}}}{_k}\}}_{k=1}^{K}$, ${\{\dot{\boldsymbol{\mathrm{Y}}}{_{1,k}}\}}_{k=1}^{K}$ and ${\{\dot{\boldsymbol{\mathrm{Y}}}{_{2,k}}\}}_{k=1}^{K}$. Let $\dot{\boldsymbol{\mathrm{Q}}}=\dot{\boldsymbol{\mathrm{J}}}{_k} -\frac{\dot{\boldsymbol{\mathrm{Y}}}{_{1,k}}}{\mu}$. The $\dot{\boldsymbol{\mathrm{Z}}}{_k}$-subproblem is reduced to:
\begin{equation}
    \begin{aligned}
        \mathop{\arg\min}_{\dot{\boldsymbol{\mathrm{Z}}}{_k}}\ \ \frac{\mu}{2}{\|\dot{\boldsymbol{\mathrm{Z}}}{_k}-\dot{\boldsymbol{\mathrm{Q}}}\|}_{F}^2+ \frac{\mu}{2}\|\mathcal{R}(\dot{\boldsymbol{\mathrm{B}}}){_k}-\dot{\boldsymbol{\mathrm{A}}}\dot{\boldsymbol{\mathrm{Z}}}{_k}+ \frac{\dot{\boldsymbol{\mathrm{Y}}}{_{2,k}}}{\mu} \|_F^2.  
    \end{aligned}\label{eq21}
\end{equation}
$\dot{\boldsymbol{\mathrm{Z}}}{_k}$ is explicitly updated as follows.
\begin{equation}
    \begin{split}
        \dot{\boldsymbol{\mathrm{Z}}}{_k}={(\dot{\boldsymbol{\mathrm{A}}}{^{\mathrm{H}}}\dot{\boldsymbol{\mathrm{A}}}+\dot{\boldsymbol{\mathrm{I}}}_{d})}^{-1}(\dot{\boldsymbol{\mathrm{A}}}{^{\mathrm{H}}}({\mathcal{R}(\dot{\boldsymbol{\mathrm{B}}})}{_k}+\frac{\dot{\boldsymbol{\mathrm{Y}}}{_{2,k}}}{\mu})+\dot{\boldsymbol{\mathrm{Q}}}).
    \end{split}
    \label{eq22}
\end{equation}

\textbf{Update $\dot{\boldsymbol{\mathrm{G}}}{_1}$ and $\dot{\boldsymbol{\mathrm{G}}}{_2}$.}  Fix $\dot{\boldsymbol{\mathrm{B}}}$ and $\dot{\boldsymbol{\mathrm{Y}}}{_{3}}$. $\dot{\boldsymbol{\mathrm{G}}}{_1}$ is updated according to the subproblem:
\begin{align}
\begin{aligned}
\dot{\boldsymbol{\mathrm{G}}}{_1}=\mathop{\arg\min}_{\dot{\boldsymbol{\mathrm{G}}}{_1}}\ \ \frac{\alpha}{\mu}{\| {\dot{\boldsymbol{\mathrm{G}}}}_1 \|}_{1}
    +\frac{1}{2}\|\dot{\boldsymbol{\mathrm{G}}}{_1}-\dot{\boldsymbol{\mathrm{M}}}_1 \|{_F^2}
\end{aligned} \label{eq23}
\end{align}
The solution to Eq. \eqref{eq23} is obtained using the soft-thresholding method in Lemma \ref{lma2} in Section~\ref{overview}. Similarly, $\dot{\boldsymbol{\mathrm{G}}}{_2}$ is updated in the same way as the $\dot{\boldsymbol{\mathrm{G}}}{_1}$-subproblem.

\textbf{Update $\dot{\boldsymbol{\mathrm{B}}}$.}  To compute the solution of $\dot{\boldsymbol{\mathrm{B}}}$-subproblem, variables ${\{\dot{\boldsymbol{\mathrm{Z}}}{_k}\}}_{k=1}^{K}$, $\dot{\boldsymbol{\mathrm{G}}}_1$, $\dot{\boldsymbol{\mathrm{G}}}_2$, $\dot{\boldsymbol{\mathrm{D}}}$, $\dot{\boldsymbol{\mathrm{E}}}$, ${\{\dot{\boldsymbol{\mathrm{Y}}}{_{2,k}}\}}_{k=1}^{K}$, $\dot{\boldsymbol{\mathrm{Y}}}_3$, $\dot{\boldsymbol{\mathrm{Y}}}_4$, and $\dot{\boldsymbol{\mathrm{Y}}}_5$ are fixed. Let $\dot{\boldsymbol{\mathrm{M}}}{_{3,k}}=\dot{\boldsymbol{\mathrm{A}}}\dot{\boldsymbol{\mathrm{Z}}}{_k}-\frac{\dot{\boldsymbol{\mathrm{Y}}}{_{2,k}}}{\mu}$. ${\{\dot{\boldsymbol{\mathrm{M}}}{_{3,k}}\}}_{k=1}^K$ is computed for each group. We reverse the extraction process of local quaternion image patches using the inverse operator ${\mathcal{R}}^{-1}$ applied to $\dot{\boldsymbol{\mathrm{M}}}{_{3}}$. It is the stacked version of ${\{\dot{\boldsymbol{\mathrm{M}}}{_{3,k}}\}}_{k=1}^K$.
Let $\dot{\boldsymbol{\mathrm{M}}}_4 =\dot{\boldsymbol{\mathrm{G}}}{_1}+\frac{\dot{\boldsymbol{\mathrm{Y}}}{_3}}{\mu}$, $\dot{\boldsymbol{\mathrm{M}}}_5=\dot{\boldsymbol{\mathrm{G}}}{_2}+\frac{\dot{\boldsymbol{\mathrm{Y}}}{_4}}{\mu}$, and $\dot{\boldsymbol{\mathrm{M}}}_6=\dot{\boldsymbol{\mathrm{I}}}-\dot{\boldsymbol{\mathrm{D}}}-\dot{\boldsymbol{\mathrm{E}}}+\frac{\dot{\boldsymbol{\mathrm{Y}}}{_5}}{\mu}$. The $\dot{\boldsymbol{\mathrm{B}}}$-subproblem can be rewritten as below.
\begin{equation}
    \begin{aligned}
    \dot{\boldsymbol{\mathrm{B}}}=&\mathop{\arg\min}_{\dot{\boldsymbol{\mathrm{B}}}}\ \ \frac{\mu}{2}\|\dot{\boldsymbol{\mathrm{B}}}- \mathcal{R}^{-1}({\dot{\boldsymbol{\mathrm{M}}}_3})\|_F^2+\frac{\mu}{2}\|\nabla_1 \dot{\boldsymbol{\mathrm{B}}}-\dot{\boldsymbol{\mathrm{M}}}_4\|{_F^2}\\
        &+\frac{\mu}{2}\|\nabla_2 \dot{\boldsymbol{\mathrm{B}}}-\dot{\boldsymbol{\mathrm{M}}}_5 \|{_F^2} 
        +\frac{\mu}{2}\|\dot{\boldsymbol{\mathrm{B}}}-\dot{\boldsymbol{\mathrm{M}}}_6\|{_F^2}.
    \end{aligned}
    \label{eq24}
\end{equation}
We set the derivatives of Eq. \eqref{eq24} with respect to $\dot{\boldsymbol{\mathrm{B}}}$ to zero. $\dot{\boldsymbol{\mathrm{B}}}$ is updated using quaternion fast Fourier transform $\mathcal{F}(\cdot)$ under the quaternion periodic boundary condition \cite{huang2022quaternion}. Let $\dot{\boldsymbol{\mathrm{\Sigma}}}$ denote the result of $\mathcal{F}({\mathcal{R}}^{-1}({\dot{\boldsymbol{\mathrm{M}}}_3}))+\mathcal{F}(\nabla_1^{\mathrm{T}})\cdot \mathcal{F}(\dot{\boldsymbol{\mathrm{M}}}_4)+\mathcal{F}(\nabla_2^{\mathrm{T}})\cdot \mathcal{F}(\dot{\boldsymbol{\mathrm{M}}}_5)+\mathcal{F}(\dot{\boldsymbol{\mathrm{M}}}_6)$. $\dot{\boldsymbol{\mathrm{B}}}$ is updated as below.
\begin{equation}
    \begin{split}
        \dot{\boldsymbol{\mathrm{B}}}=\mathcal{F}{^{-1}}({\frac{\dot{\boldsymbol{\mathrm{\Sigma}}}}{ \mathcal{F}(\nabla_1^{\mathrm{T}}\nabla_1+\nabla_2^{\mathrm{T}}\nabla_2)+2}}).
    \end{split}\label{eq25}
\end{equation}

\textbf{Update $\dot{\boldsymbol{\mathrm{D}}}$.} We fix $\dot{\boldsymbol{\mathrm{B}}}$, $\dot{\boldsymbol{\mathrm{E}}}$, and $\dot{\boldsymbol{\mathrm{Y}}}_4$ to minimize $\dot{\boldsymbol{\mathrm{D}}}$. The $\dot{\boldsymbol{\mathrm{D}}}$-subproblem can be rewritten as follows:
\begin{equation}
    \begin{split}
       \dot{\boldsymbol{\mathrm{D}}}=\mathop{\arg\min}_{\dot{\boldsymbol{\mathrm{D}}}}\ \ \frac{\beta}{\mu} {\|\dot{\boldsymbol{\mathrm{D}}} \|}_1 +\frac{1}{2}\| \dot{\boldsymbol{\mathrm{D}}}-\dot{\boldsymbol{\mathrm{M}}}_7 \|_F^2.
    \end{split}
    \label{eq26}
\end{equation}
Eq. \eqref{eq26} can also be solved using Lemma \ref{lma2} in Section~\ref{overview}.

\textbf{Update $\dot{\boldsymbol{\mathrm{E}}}$.}  Fix $\dot{\boldsymbol{\mathrm{B}}}$, $\dot{\boldsymbol{\mathrm{D}}}$, and $\dot{\boldsymbol{\mathrm{Y}}}_5$. Let $\dot{\boldsymbol{\mathrm{M}}}_8=\dot{\boldsymbol{\mathrm{I}}}-\dot{\boldsymbol{\mathrm{D}}}-\dot{\boldsymbol{\mathrm{B}}}+\frac{\dot{\boldsymbol{\mathrm{Y}}}{_5}}{\mu}$. The solution of the $E$-subproblem is obtained by setting the gradient values of $\dot{\boldsymbol{\mathrm{E}}}$'s subproblem  with respect to $\dot{\boldsymbol{\mathrm{E}}}$ to zero. $\dot{\boldsymbol{\mathrm{E}}}$ is updated as below.
\begin{equation}
    \dot{\boldsymbol{\mathrm{E}}}={(2\lambda+\mu \dot{\boldsymbol{\mathrm{I}}}_{d})}^{-1}(\mu \dot{\boldsymbol{\mathrm{M}}}_8),
    \label{eq27}
\end{equation}

The optimization process of the QFED model is shown in Algorithm \ref{algorithm1}.
\begin{algorithm}
    \caption{Quaternion focal element decomposition (QFED)}\label{algorithm1}
    \KwIn{ The quaternion representation of the source image $\dot{\boldsymbol{\mathrm{I}}}$, the parameters $\mu$, $\lambda$, $\alpha$ and $\beta$, the iteration number $t$.}
    \KwOut{Optimal quaternion coefficient matrix $\dot{\boldsymbol{\mathrm{Z}}}$ and detail layer $\dot{\boldsymbol{\mathrm{D}}}$}
    Initialize ${\{\dot{\boldsymbol{\mathrm{J}}}{_k^0}\}}{_{k=1}^K}$,${\{\dot{\boldsymbol{\mathrm{Z}}}{_k^0}\}}{_{k=1}^K}$, ${\{\dot{\boldsymbol{\mathrm{Y}}}{_{1,k}^0}\}}{_{k=1}^K}$,${\{\dot{\boldsymbol{\mathrm{Y}}}{_{2,k}^0}\}}{_{k=1}^K}$, $\dot{\boldsymbol{\mathrm{G}}}{_1^0}$, $\dot{\boldsymbol{\mathrm{G}}}{_2^0}$, $\dot{\boldsymbol{\mathrm{B}}}{^0}$, $\dot{\boldsymbol{\mathrm{D}}}{^0}$, $\dot{\boldsymbol{\mathrm{E}}}{^0}$, $\dot{\boldsymbol{\mathrm{Y}}}{_3^0}$, $\dot{\boldsymbol{\mathrm{Y}}}{_4^0}$ and $\dot{\boldsymbol{\mathrm{Y}}}{_5^0}$;
    
    \While{\textnormal{not converged}}{
    \For{$k=1,\cdots,K$}{
     Fix other variables and compute $\dot{\boldsymbol{\mathrm{J}}}{_k^t}$ using Lemma \ref{lma1};
     
     Fix other variables and compute $\dot{\boldsymbol{\mathrm{Z}}}{_k^t}$ using Eq. \eqref{eq22};
    }
    Fix other variables and solve $\dot{\boldsymbol{\mathrm{G}}}{_1^t}$ and $\dot{\boldsymbol{\mathrm{G}}}{_2^t}$ using Lemma \ref{lma2} respectively;

    Fix other variables and solve $\dot{\boldsymbol{\mathrm{B}}}{^t}$ using Eq. \eqref{eq25};

    Fix other variables and solve $\dot{\boldsymbol{\mathrm{D}}}{^t}$ using Lemma \ref{lma2};

    Fix other variables and solve $\dot{\boldsymbol{\mathrm{E}}}{^t}$ using Eq. \eqref{eq27};

    \For{$k=1,\cdots,K$}{
    $\dot{\boldsymbol{\mathrm{Y}}}{_{1,k}^t}=\dot{\boldsymbol{\mathrm{Y}}}{_{1,k}^{t-1}}+\mu(\dot{\boldsymbol{\mathrm{Z}}}{_k^t}-\dot{\boldsymbol{\mathrm{J}}}{_k^t})$;

     $\dot{\boldsymbol{\mathrm{Y}}}{_{2,k}^t}=\dot{\boldsymbol{\mathrm{Y}}}{_{2,k}^{t-1}}+\mu(\mathcal{R}(\dot{\boldsymbol{\mathrm{B}}}){_k^{t}}-\dot{\boldsymbol{\mathrm{A}}}\dot{\boldsymbol{\mathrm{Z}}}{_k^{t}})$;
    }
     
     ${\dot{\boldsymbol{\mathrm{Y}}}{_{3}^t}}={\dot{\boldsymbol{\mathrm{Y}}}{_{3}^{t-1}}}+\mu(\dot{\boldsymbol{\mathrm{G}}}{_1^{t}}-\nabla_1 \dot{\boldsymbol{\mathrm{B}}}{^t})$;

     ${\dot{\boldsymbol{\mathrm{Y}}}{_{4}^t}}={\dot{\boldsymbol{\mathrm{Y}}}{_{4}^{t-1}}}+\mu(\dot{\boldsymbol{\mathrm{G}}}{_2^{t}}-\nabla_2 \dot{\boldsymbol{\mathrm{B}}}{^t})$;

     ${\dot{\boldsymbol{\mathrm{Y}}}{_{5}^t}}={\dot{\boldsymbol{\mathrm{Y}}}{_{5}^{t-1}}}+\mu(\dot{\boldsymbol{\mathrm{I}}}-\dot{\boldsymbol{\mathrm{B}}}{^t}-\dot{\boldsymbol{\mathrm{D}}}{^t}-\dot{\boldsymbol{\mathrm{E}}}{^t})$;

       $\mu=\min{\{{10}^6,\mu*1.1\}}$;

       $t$ $\longleftarrow$ $t+1$;
    }
\end{algorithm}
\subsubsection{Patch-wise dual-scale focus map generation}

As illustrated in Fig. \ref{fig:QCAFD-flowchart}, we obtain the corresponding detail layers $\dot{\boldsymbol{\mathrm{D}}}{_1}$, $\dot{\boldsymbol{\mathrm{D}}}{_2}$ and coefficient matrices $\dot{\boldsymbol{\mathrm{Z}}}{_1}$, $\dot{\boldsymbol{\mathrm{Z}}}{_2}$ from the QFED process for given input quaternion representations $\dot{\boldsymbol{\mathrm{I}}}{_1}$, $\dot{\boldsymbol{\mathrm{I}}}{_2}$. To fully leverage the complementary information in $\dot{\boldsymbol{\mathrm{D}}}$ and $\dot{\boldsymbol{\mathrm{Z}}}$, we propose a patch-wise dual-scale focus map generation strategy to generate the base-scale and detail-scale focus maps at the patch level. 

\textbf{Detail Amplification}. The detail-scale layer for each pixel location is amplified within a sliding window to strengthen the fine-grained details:
\begin{equation}
    \dot{\boldsymbol{\mathrm{D}}}{_s}(x,y)=\sum_{s=-r}^{r}\sum_{t=-r}^{r}\dot{\boldsymbol{\mathrm{D}}}(x+s,y+t), \label{eq:amplification}
\end{equation}
where $\dot{\boldsymbol{\mathrm{D}}}{_s}(x,y)$ represents the sum of all pixels within the window centered on the pixel $\dot{\boldsymbol{\mathrm{D}}}(x,y)$; the window size is $2r+1$ and $r=3$ by default.

\textbf{Dual-scale focus measure}. The dual-scale focus measures used to compute patch-wise focus levels are defined as:
\begin{align}
    & l_{B}={\| \nabla_1 \dot{\boldsymbol{\mathrm{d}}} \|}_1+{\| \nabla_2 \dot{\boldsymbol{\mathrm{d}}} \|}_1 + \theta {\|\dot{\boldsymbol{\mathrm{z}}}\| }_2, \label{eq-base-scale}\\
    & l_{D}=\phi({\| \nabla_1 \dot{\boldsymbol{\mathrm{d}}}_s \|}_1+{\| \nabla_2 \dot{\boldsymbol{\mathrm{d}}}_s \|}_1),\label{eq-detail-scale}
\end{align}
where $l_{B}$ and $l_{D}$ represent patch-wise base-scale and detail-scale focus levels of input quaternion representation $\dot{\boldsymbol{\mathrm{I}}}$; $\dot{\boldsymbol{\mathrm{d}}}$ and $\dot{\boldsymbol{\mathrm{z}}}$ denote the local patch and corresponding column vector of $\dot{\boldsymbol{\mathrm{D}}} $ and $\dot{\boldsymbol{\mathrm{Z}}} $ respectively; $\dot{\boldsymbol{\mathrm{d}}}_s $  represents the local patch of $\dot{\boldsymbol{\mathrm{D}}}_s $ computed by Eq. \eqref{eq:amplification}. Parameter $\theta$ is set to 1 by default. $\phi(\cdot)$ is a Laplacian-based enhancement function defined as $\phi(x) = 1 - e^{-x / \gamma}$ where $\gamma=0.2$ by default.
The detail-scale focus measure captures the magnitude of two-directional gradients that effectively reflect focus levels in color images. The base-scale focus measure jointly considers gradient magnitudes and coefficient matrix energy, thereby improving spatial coherence and mitigating local artifacts.

To compare base-scale and detail-scale focus levels of the local patches in $\dot{\boldsymbol{\mathrm{I}}}{_1}$ and $\dot{\boldsymbol{\mathrm{I}}}{_2}$, the patch-wise dual-scale focus maps $\boldsymbol{\mathrm{m}}{_B}$ and $\boldsymbol{\mathrm{m}}{_D}$ are computed as:
\begin{align}
&\boldsymbol{\mathrm{m}}{_B}=\begin{cases}
        0, & {\rm if}\   l_{B,1}>l_{B,2} \\
        1, & {\rm if} \  l_{B,1}\leq l_{B,2}
    \end{cases} \label{eq-comparison-base}\\
    &\boldsymbol{\mathrm{m}}{_D}=\begin{cases}
        0, & {\rm if}\ l_{D,1}>l_{D,2} \\
        1, & {\rm if}\ l_{D,1}\leq l_{D,2}
    \end{cases}
    \label{eq-comparison-detail}
\end{align}
where $l_{B,1}$ and $l_{B,2}$ are computed using Eq. \eqref{eq-base-scale}; $l_{D,1}$ and $l_{D,2}$ are computed using Eq. \eqref{eq-detail-scale}; $\boldsymbol{\mathrm{m}}{_B}$ and $\boldsymbol{\mathrm{m}}{_D}$ denote patch-wise base-scale and detail-scale focus maps respectively.
Aggregating these decisions across all patches yields the global base-scale and detail-scale focus maps $\boldsymbol{\mathrm{M}}{_B}$ and $\boldsymbol{\mathrm{M}}{_D}$.

\begin{figure}[htbp]
\centering {\includegraphics[width=1\columnwidth]{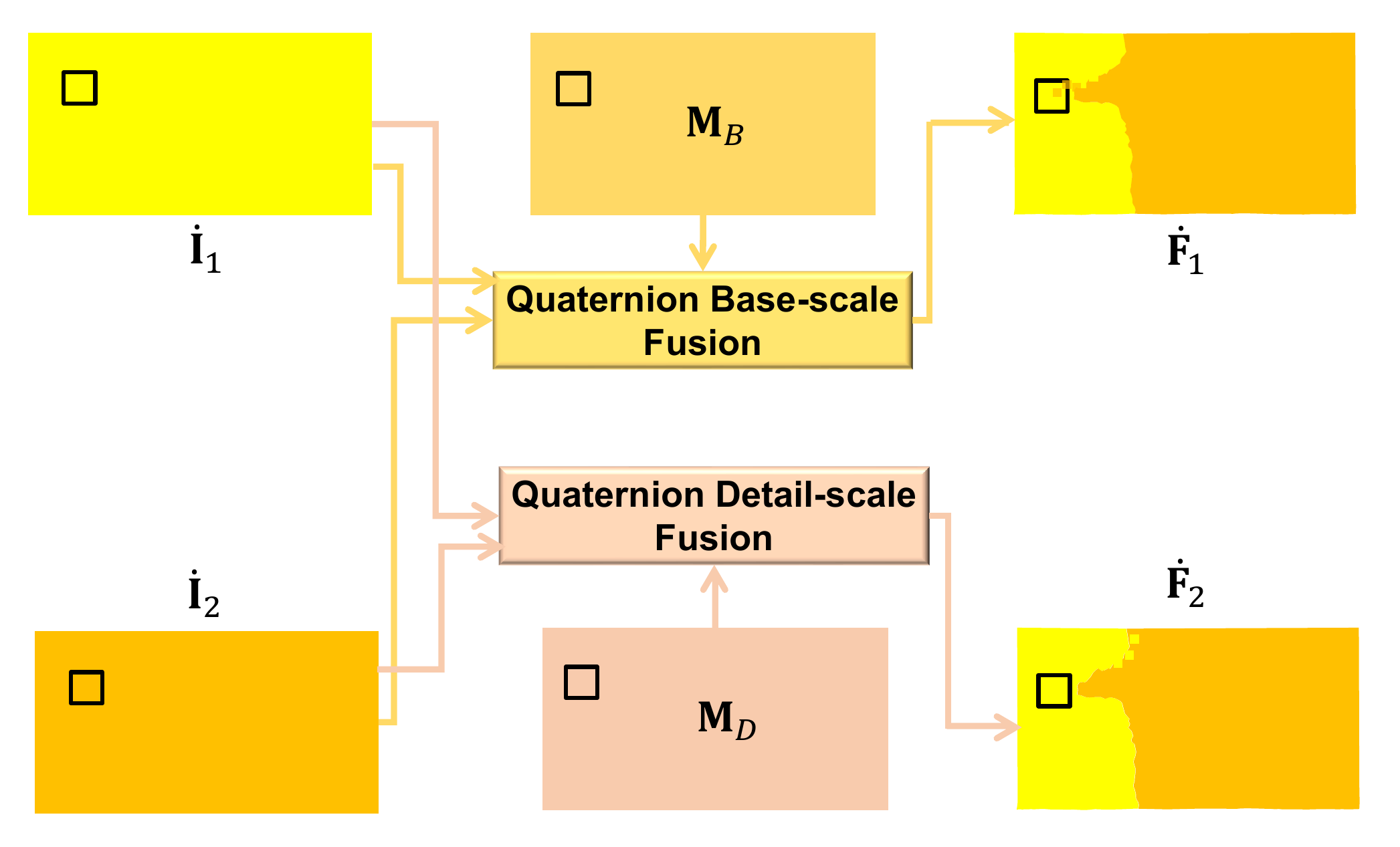}}
\caption{Our quaternion base-detail fusion strategy.} \label{fig:QBDF-flowchart}
\end{figure}
\subsection{Quaternion base-detail fusion strategy} \label{fusion}
Our quaternion base-detail fusion (QBDF) strategy aims to achieve high-quality multi-focus color image fusion that operates at base and detail scales. The overall fusion process is illustrated in Fig. \ref{fig:QBDF-flowchart}. Given the input quaternion representations $\dot{\boldsymbol{\mathrm{I}}}_{1}$ $\in \mathbbm{H}^{M \times N}$ and $\dot{\boldsymbol{\mathrm{I}}}_{2}$ $\in \mathbbm{H}^{M \times N}$ and corresponding dual-scale focus maps $\boldsymbol{\mathrm{M}}_{B}$ and $\boldsymbol{\mathrm{M}}_{D}$, the QBDF strategy performs dual-scale image fusion.

The patch-wise selection results that are derived from the focus maps are used to construct the fused images at each scale. Specifically, the base-scale fused result ${\dot{\boldsymbol{\mathrm{F}}}}_{1}$ and detail-scale fused result ${\dot{\boldsymbol{\mathrm{F}}}}_{2}$ are computed using the following rules:
 \begin{align}        
    &{\dot{\boldsymbol{\mathrm{F}}}}_{1}=\left \{ 
    \begin{aligned}
        &{\dot{\boldsymbol{\mathrm{I}}}}_{1},& & {\rm if} \  \boldsymbol{\mathrm{M}}_{B}=0\\
        &{\dot{\boldsymbol{\mathrm{I}}}}_{2},& & {\rm if} \  \boldsymbol{\mathrm{M}}_{B}=1
    \end{aligned}
    \right. \label{p1}\\
    &{\dot{\boldsymbol{\mathrm{F}}}}_{2}=\left \{ 
    \begin{aligned}
        &{\dot{\boldsymbol{\mathrm{I}}}}_{1},& & {\rm if} \  \boldsymbol{\mathrm{M}}_{D}=0\\
        &{\dot{\boldsymbol{\mathrm{I}}}}_{2},& & {\rm if} \  \boldsymbol{\mathrm{M}}_{D}=1
    \end{aligned}
    \right.\label{p2}
    \end{align}
This fusion rule ensures that the source image with the most focused content is selected and integrated at detail or base scales for each patch location.
\subsection{Quaternion structural similarity refinement strategy} \label{refinement}
To further enhance the quality of the final fused image, we propose a quaternion structural similarity refinement (QSSR) strategy. It refines the local patches of base-scale and detail-scale fusion results $ \dot{\boldsymbol{\mathrm{F}}}_{1}$ and $ \dot{\boldsymbol{\mathrm{F}}}_{2}$ using a patch-wise selection rule. The assumption is that the optimal fused patch should exhibit the highest similarity with the corresponding region in the source image that contains the most focused content. We first propose a weighted quaternion structure similarity ($WQ_{SSIM}$) measure.

\textbf{Weighted quaternion structural similarity measure}.
For each spatial location, let $ \dot{\boldsymbol{\mathrm{f}}}$ denote a local patch of a base-scale or detail-scale fusion result $\dot{\boldsymbol{\mathrm{F}}}_{2}$ or $\dot{\boldsymbol{\mathrm{F}}}_{2}$. $\dot{\boldsymbol{\mathrm{p}}}_{1}$ and $\dot{\boldsymbol{\mathrm{p}}}_{2}$ represent the corresponding local patches extracted from $\dot{\boldsymbol{\mathrm{I}}}_{1}$ and $\dot{\boldsymbol{\mathrm{I}}}_{2}$ at the same spatial location. 
The $WQ_{SSIM}$ value is computed as follows:
\begin{equation}
\begin{aligned}
&WQ_{SSIM}({\dot{\boldsymbol{\mathrm{f}}}})= {\tau}_{1} Q_{SSIM}(\dot{\boldsymbol{\mathrm{f}}},\dot{\boldsymbol{\mathrm{p}}}_1)+{\tau}_{2} Q_{SSIM}(\dot{\boldsymbol{\mathrm{f}}},\dot{\boldsymbol{\mathrm{p}}}_2)
\end{aligned}
\label{QWSSIM}
\end{equation}
where quaternion structural similarity ($Q_{SSIM}$) is defined in Definition \ref{dfn5} in Section \ref{overview};
${\tau}_1$ and ${\tau}_2$ are the adaptive weights assigned to $\dot{\boldsymbol{\mathrm{p}}}_1$ and $\dot{\boldsymbol{\mathrm{p}}}_2$ respectively based on its detail-scale focus level. 
They are defined below:
\begin{equation}
    {\tau}_{1}=\frac{l_{D,1}}{l_{D,1}+l_{D,2}+\epsilon}, \quad {\tau}_{2}=1-{\tau}_{1}
    \label{adaptiveweights}
\end{equation}
where $\epsilon$ is a small positive constant.

$WQ_{SSIM}$ guides the selection of the optimal patch at each spatial location.
The final fused patch $\dot{\boldsymbol{\mathrm{p}}}$ at each location is selected according to:
\begin{equation}
    \dot{\boldsymbol{\mathrm{p}}}=\left \{
    \begin{aligned}
        &\dot{\boldsymbol{\mathrm{f}}}_1, &{\rm if} \  WQ_{SSIM}({\dot{\boldsymbol{\mathrm{f}}}}_{1})>WQ_{SSIM}({\dot{\boldsymbol{\mathrm{f}}}}_{2})\\
        &\dot{\boldsymbol{\mathrm{f}}}_2, &{\rm if} \  WQ_{SSIM}({\dot{\boldsymbol{\mathrm{f}}}}_{1})\leq WQ_{SSIM}({\dot{\boldsymbol{\mathrm{f}}}}_{2})\end{aligned} 
    \right.\label{eq-MS}
\end{equation}
This strategy aims to effectively suppress artifacts and corrects misclassified regions in the fused outputs. Aggregating all selected patches according to Eq. \eqref{eq-MS} yields the quaternion fused result $\dot{\boldsymbol{\mathrm{F}}}$. Convert $\dot{\boldsymbol{\mathrm{F}}}$ back to the real domain to obtain the fused color image $\boldsymbol{\mathrm{F}}$.

\section{Experiments} \label{experiments}
To evaluate the fusion performance of our QMCIF framework, we conduct experiments on three public datasets. Section A presents experimental settings like competing methods, datasets, and evaluation metrics. Section B properties of the quaternion focal element decomposition (QFED) module that is a key component of the quaternion consistency-aware focus detection (QCAFD) method. 
To demonstrate the superiority of our QMCIF framework,  Section C presents experimental results compared with state-of-the-art approaches. Section D presents ablation studies on the mffw dataset \cite{xu2020mffw} to verify the effectiveness of our QCAFD module, QBDF and QSSR strategies.
\subsection{ Experimental settings} \label{ex-setting}
In our experiments, the patch size is set as $\sqrt{d}\times\sqrt{d}$ where $\sqrt{d}=8$ by default in the QCAFD method and base-scale fusion process. In the detail-scale fusion and QSSR process, $\sqrt{d}$ is set to $5\times10^{-5}MN$ \cite{xu2020towards}, where $M$ and $N$ denote the height and width of the input image. The stopping criterion of QFED is $1e-5$ in Algorithm 1. All experiments are performed using Matlab 2016 on a workstation with a 2.90GHz Intel Core CPU and 16GB memory.

\textbf{Competing methods}. Eleven representative multi-focus image fusion algorithms are selected as the comparison methods. They include a sparse coding-based (SR) method \cite{yang2009multifocus}, a convolutional sparse coding-bsed (CSR) method \cite{liu2016image}, a multi-scale gradients-based (MGIMF)  method \cite{chen2021multi}, GAN-based (MFIF-GAN) method \cite{wang2021mfif}, a higher-order SVD-based (QHOSVD)  method \cite{miao2023quaternion}, unsupervised zero-shot based method (ZMFF) \cite{hu2023zmff}, a small-area-aware (SAMF) method \cite{li2024samf}, End-to-End based network (DBMFIF) \cite{zhang2024exploit}, General-Image-Fusion-based (TCMOA) \cite{zhu2024task}, a Swintransformer-based network (SwinMFF) \cite{xie2024swinmff} and a convolutional sparse representation-based unfolding network (MCCSR-Net) \cite{zheng2025CCSR}. SR and CSR are channel-wise processing-based methods. MGIMF and SAMF are grayscale conversion-based methods. QHOSVD is a quaternion domain-based method. MFIF-GAN, ZMFF, DBMFIF, TCMOA, SwinMFF and MCCSR-Net are deep-learning-based methods. TCMOA and SwinMFF are cross-channel attention-based methods.

\textbf{Datasets}. In our experiments, we utilize three public datasets without ground truth: the lytro dataset \cite{nejati2015multi}, mffw dataset \cite{xu2020mffw}, and MFI-WHU dataset \cite{zhang2021mff}. The lytro dataset contains 20 pairs of samples for two-color image fusion and 4 pairs of samples for multiple-color image fusion. The mffw dataset includes 13 pairs of samples for two-color image fusion and 6 pairs of image data for multiple-source fusion. The MFI-WHU dataset provides 30 pairs of samples for two-color image fusion.

The images in the lytro dataset \cite{nejati2015multi} are well-registered and exhibit high levels of details. It aims to evaluate focus information preservation in ideal conditions. The images in the mffw dataset \cite{xu2020mffw} are affected by severe defocus spread effects with large misalignment and contain both low-detail and high-detail regions. The performance on the mffw dataset provides a robust evaluation of focus information preservation and artifact removal in complex real-world scenarios. The MFI-WHU dataset \cite{zhang2021mff} includes input images with tiny blurred objects. These pose challenges for small-area focus detection and fusion.

\textbf{Metrics}. Following \cite{liu2020multi}, six objective metrics are adopted to evaluate the performance of color image fusion quantitatively. They are the normalized mutual information $Q_{MI}$, gradient-based metric $Q_G$, phase congruency-based metric $Q_P$, structural similarity-based metric $Q_E$, $Q_Y$ and human perception-based metric $Q_{CB}$. $Q_{MI}$ quantifies the intensity-based mutual information transferred from source images to the fused output. $Q_G$ and $Q_P$ assess the preservation of gradient information and salient feature details. $Q_Y$ evaluates the structural preservation of the fused image in terms of luminance, contrast, and texture patterns while $Q_E$ extends $Q_Y$ by considering edge consistency in the fused result. $Q_{CB}$ measures the preservation of contrast variations from source images in the fused image.

\subsection{Analysis of QFED} \label{QFED}
This subsection analyzes our quaternion focal element decomposition (QFED) model that is a key component of QCAFD. We aim to validate QFED's convergence, parameter sensitivity, and the impact of its each term on the whole fusion performance.

\begin{figure}[hbtp]
\centering \includegraphics[width=0.6\columnwidth]{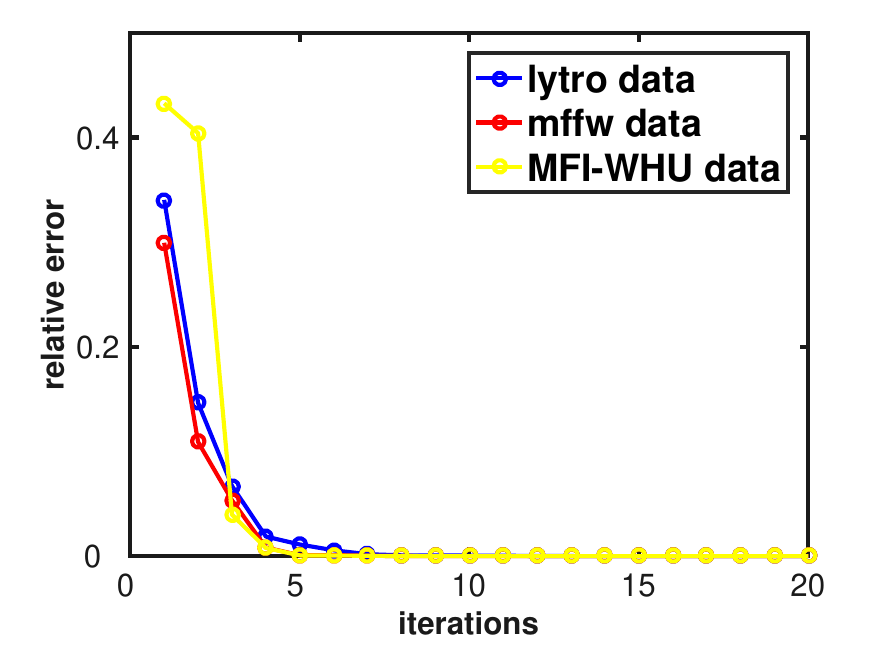} 
\caption{Convergence analysis of QFED model on lytro, mffw and MFI-WHU datasets.} \label{fig:Convergence analysis} \end{figure}

\textbf{Convergence Analysis}. 
Fig. \ref{fig:Convergence analysis} displays the evolution curves of the relative difference versus iterations of QFED on
the samples of lytro \cite{nejati2015multi}, mffw \cite{xu2020mffw} and MFI-WHU datasets \cite{zhang2021mff}. The relative difference is defined by the maximum value between $\| \dot{\boldsymbol{\mathrm{Z}}}{^{t+1}}-\dot{\boldsymbol{\mathrm{Z}}}{^t}\|{_{\infty}}$ and $\| \dot{\boldsymbol{\mathrm{D}}}{^{t+1}}-\dot{\boldsymbol{\mathrm{D}}}{^{t}}\|{_{\infty}}$. $\dot{\boldsymbol{\mathrm{Z}}}{^{t+1}}$ and $\dot{\boldsymbol{\mathrm{Z}}}{^{t}}$ denote the coefficient matrices at successive $t^{th}$ and ${(t+1)}^{th}$ iterations respectively. $\dot{\boldsymbol{\mathrm{D}}}{^{t+1}}$ and $\dot{\boldsymbol{\mathrm{D}}}{^{t}}$ denote the detail layers at successive $t^{th}$ and ${(t+1)}^{th}$ iterations respectively. The $\infty$ norm means the largest magnitude in each element of a vector. To improve the QFED speed, $\dot{\boldsymbol{\mathrm{B}}}{^{0}}$ is initialized as $\frac{\mathcal{F}(\dot{\boldsymbol{\mathrm{I}}})}{ \mathcal{F}(\nabla_1^T\nabla_1+\nabla_2^T\nabla_2)+1}$ and $\dot{\boldsymbol{\mathrm{D}}}{^{0}}$ is set to $\dot{\boldsymbol{\mathrm{I}}}-\dot{\boldsymbol{\mathrm{B}}}{^{0}}$ .
As displayed in Fig. \ref{fig:Convergence analysis}, QFED is converges within 20 iterations. This demonstrates the computational efficiency and stability of the iterative solution of QFED.

\textbf{Parameter Sensitivity}.
We analyze the influence of regularization parameters $\alpha$, $\beta$, and $\lambda$ in the QFED model in Eq.~\eqref{eq6}.
\begin{figure}[hbtp]
\centering \includegraphics[width=\columnwidth]{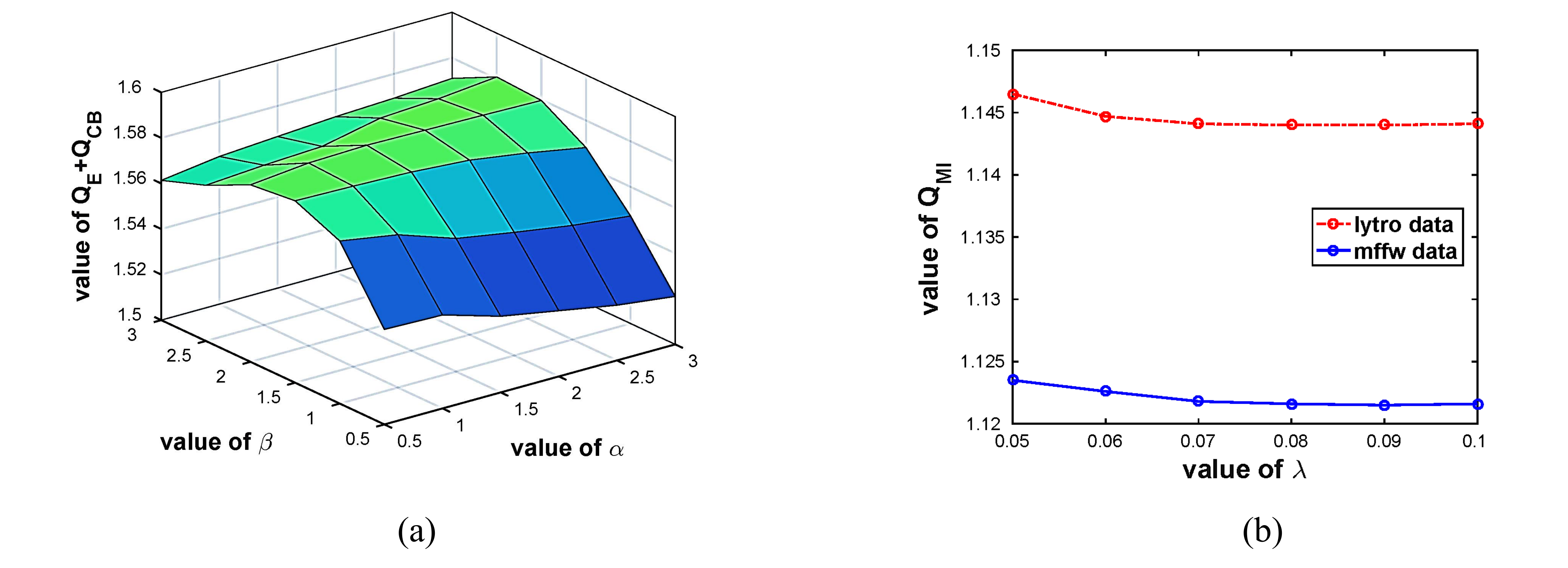} 
\caption{Fusion performance with different parameters on the lytro and the mffw datasets. 
(a) $Q_{E}+Q_{CB}$ results when parameters $\alpha$ and $\beta$ vary on the mffw dataset. (b) The $Q_{MI}$ value with respect to parameter $\lambda$ on both lytro and mffw datasets.} \label{fig:paras4} 
\end{figure}
In QFED, parameter $\alpha$ controls the base-scale layer while parameter $\beta$ handles the detail-scale layer. The final fused image not only preserves the most detailed regions of the inputs but also maintains the consistency of the background. 

To handle the three regularization parameters in our QFED model, we fix parameter $\lambda=0.05$ and adjust the others each time. $\alpha$ and $\beta$ are sequentially set values of [0.5,
1, 1.5, 2, 2.5, 3]. Then $\alpha$ and $\beta$
are set to 1.5 and 2 and the variation values of $\lambda$ are sequentially
set values of [0.05, 0.06, 0.07, 0.08, 0.09, 0.1]. To effectively estimate the fusion performance, the combined quality metric $Q_{E}+Q_{CB}$ jointly assesses structural consistency and perceptual quality. Since $\lambda$ is not remarkably sensitive to the final fusion quality, $Q_{MI}$ is used to quantitatively analyze the fused results. We select the values corresponding to the maximum quality measure for the balance between detail preservation and background consistency. 

 Fig. \ref{fig:paras4} (a) displays the $Q_{E}+Q_{CB}$ results of QFED
with the samples of mffw dataset. $\alpha$ and $\beta$ are set to 1.5 and 2 on mffw dataset as this setting yields the highest combined quality score. For lytro and MFI-WHU datasets, we select the optimal parameters $\alpha$ and $\beta$ in this manner and they are set to 1.5 and 0.5. In Fig. \ref{fig:paras4} (b), $Q_{MI}$ achieves its peak performance at $\lambda=0.05$. We adopt $\lambda=0.05$ consistently for all datasets since $\lambda$ exhibits less sensitivity to fusion quality variations. The parameter settings are uniformly employed in all subsequent experiments and evaluations.
\begin{table}[hbtp]
\caption{ Ablation study on each term of QFED. The best results are \textbf{bolded}.}
    \centering \resizebox{1\columnwidth}{!}{
    \begin{tabular}{|c|c|c|c|c|c|c|}
    \hline
          Term settings&$Q_{MI} \uparrow$  &$Q_G \uparrow$  & $Q_P \uparrow$  & $Q_E \uparrow$ & $Q_{CB} \uparrow$ \\
         \hline
         \textbf{$\boldsymbol{\alpha=0} $} &1.0580 &0.7359 &0.7488 &0.8240 &0.7298 \\ 
         \textbf{$\boldsymbol{\beta=0}$} &1.0052 &0.7338 &0.7282 &\textbf{0.8279} &0.6980 \\ 
         \textbf{$\boldsymbol{\lambda=0}$} &1.0140 &0.7324 &0.7432 &0.8259 &0.7127 \\ 
         \textbf{QFED} &\textbf{1.0685} &\textbf{0.7374} &\textbf{0.7832} &0.8234 &\textbf{0.7508} \\
         \hline
    \end{tabular} 
    }
    \label{tab:table99}
\end{table}

\begin{figure*}[b]
\centering
\includegraphics[width=1\textwidth]{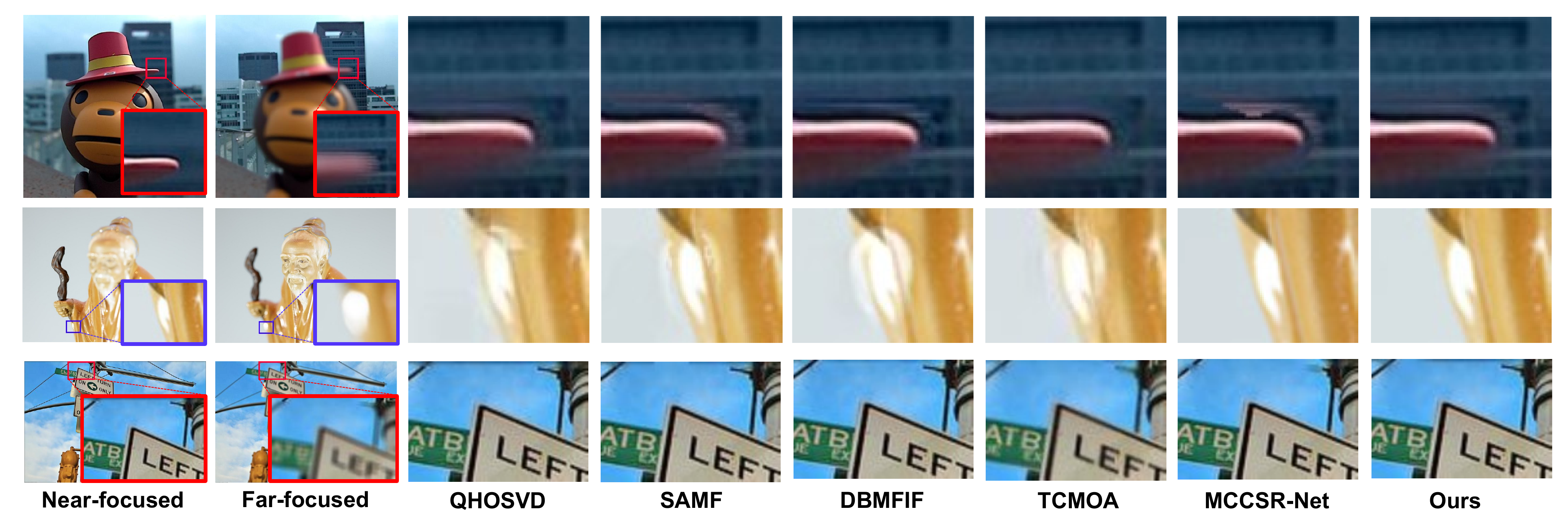} 
\caption{Two partially focused color image fusion on lytro, mffw and MFI-WHU datasets.}
\label{lytro05}
\end{figure*}
\textbf{Effectiveness of QFED.} To verify the effectiveness of the QFED model, we perform the ablation study on each term of QFED. Specifically, three different settings of $\alpha =0$, $\beta =0$, $\lambda =0$ and $\alpha \neq 0$, $\beta \neq 0$, $\lambda \neq 0$ are particularly considered in this experiment. Table \ref{tab:table99} gives the quantitative fusion results obtained by removing each term of QFED separately. As can be observed, each term can influence the fusion performance at different degrees. When $\beta$ is set to zero, we may obtain more focus-related features in the detail-scale layer $\dot{\boldsymbol{\mathrm{D}}}$. However, it fails to balance a tradeoff in the dual-scale focus map generation and subsequent fusion process. 
QFED provides the best quantitative results.

\subsection{Performance Comparison} \label{two-color-image-fusion}
This subsection discusses the quantitative and qualitative comparison results of color image fusion.

\textbf{Quantitative evaluation}. Table \ref{tab:Lytro} provides the average performance of various competing algorithms on three public datasets. Six evaluation metrics are employed to assess the quality of the fused images, with higher values indicating better performance. In this paper, the highest evaluation scores are highlighted in bold.

On the lytro dataset \cite{nejati2015multi}, our framework achieves the highest metric values compared to other competing methods across gradient-based, structural similarity-based, and human-perception-based evaluation metrics. This improvement is attributed to accurate focus detection capability of our QMCIF framework. However, our framework is worse than MCCSR-Net and SAMF on the metric of $Q_{MI}$. This is because our patch-wise fusion strategy tends to preserve less image intensity information during the quaternion fusion and refinement process compared to the pixel-wise fusion strategy.

Mffw dataset is characterized by severe defocus spread and complex low-gradient regions \cite{xu2020mffw}, all competing methods perform worse on this dataset compared to lytro and MFI-WHU datasets. Our method achieves the highest fusion performance on this dataset, significantly surpassing the second-best competing methods such as SAMF and MFIF-GAN by 1.2$\%$. In the $Q_{MI}$ results, our framework slightly outperforms SAMF thanks to
the effective artifact removal of our QSSR strategy. On the metrics of $Q_{Y}$ and $Q_{CB}$, our framework is worse only than MFIF-GAN. This demonstrates the advantages of fusion and refinement provided by the GAN-based MCIF algorithm in mitigating defocus spread effects \cite{wang2021mfif}.
\begin{table}[hbtp] 
\caption{Performance comparison of multi-focus image fusion methods.}
    \centering \resizebox{1\columnwidth}{!}{
    \begin{tabular}{|c|l|c|c|c|c|c|c|}
    \hline
         Dataset& Methods &$Q_{MI} \uparrow$ &$Q_G \uparrow$  & $Q_P \uparrow$  & $Q_E \uparrow$ & $Q_Y \uparrow$  & $Q_{CB} \uparrow$ \\
         \hline
         \multirow{9}{*}{\rotatebox{90}{\textbf{lytro \cite{nejati2015multi}}}}
         &\textbf{SR \cite{yang2009multifocus}} & 1.0642 &0.7445 &0.8168 &0.8797 &0.9688 &0.7756 \\
         &\textbf{CSR \cite{liu2016image}} & 1.0030 &0.7353 &0.8294 &0.8794 &0.9337 &0.7612 \\
         &\textbf{MGIMF \cite{chen2021multi}} & 1.1154 &0.7428 &0.8106 &0.8419 &0.9857 &0.7911 \\
         &\textbf{QHOSVD \cite{miao2023quaternion}} &1.0540	&0.7491	&0.8310	&0.8800	&0.9769	&0.7839\\
         &\textbf{SAMF \cite{li2024samf} } &1.1781		&0.7602	&0.8412	&0.8797	&0.9877	&0.8014\\
         &\textbf{MFIF-GAN \cite{wang2021mfif}} &1.0945	&0.7179	&0.8307	&0.8787	&0.9770 &0.7976\\ 
          &\textbf{ZMFF \cite{hu2023zmff} } &0.8796	&0.7013	&0.7830	&0.8678	&0.9313 &0.7399\\
         &\textbf{DBMFIF \cite{zhang2024exploit} } &1.0589	&0.7480	&0.8398	&0.8802	&0.9651 &0.7775\\
         &\textbf{TCMOA \cite{zhu2024task} } &0.9926	&0.7401	&0.8188	&0.8782	&0.9574 &0.7597\\ 
         &\textbf{SwinMFF \cite{xie2024swinmff} } &0.7689	&0.7013	&0.7773	&0.8257	&0.8867 &0.6413\\ 
         &\textbf{MCCSR-Net \cite{zheng2025CCSR} } &\textbf{1.1920}	&\textbf{0.7619}	&0.8456	&0.8772	&0.9886 &0.8084\\ 
         &\textbf{Ours} &1.1656 &0.7603 &\textbf{0.8472} &\textbf{0.8814} &\textbf{0.9896} &\textbf{0.8108} \\
         \hline
         \multirow{9}{*}{\rotatebox{90}{\textbf{mffw \cite{xu2020mffw}}}}
         &\textbf{SR \cite{yang2009multifocus}} & 0.7181 &0.6278 &0.5500 &0.8078 &0.9349 &0.6503 \\
         &\textbf{CSR \cite{liu2016image}} & 0.9023 &0.7065 &0.6903 &0.8222 &0.8556 &0.6861 \\
         &\textbf{MGIMF \cite{chen2021multi}} & 1.0529 &0.7309 &0.7363 &0.8096 &0.9507 &0.7338 \\
         &\textbf{QHOSVD \cite{miao2023quaternion}} &0.7224	&0.7249	&0.7383	&0.8187	&0.8221	&0.6050\\
         &\textbf{SAMF \cite{li2024samf} } &1.0863	&0.7310	&0.6995	&0.7974	&0.9332	&0.7101\\
         &\textbf{MFIF-GAN \cite{wang2021mfif}} &1.0681	&0.7320	&0.7550	&0.8239	&\textbf{0.9735}	&\textbf{0.7558}\\  
         &\textbf{ZMFF \cite{hu2023zmff} } &0.7728	&0.6651	&0.6476	&0.7985	&0.8775&0.6770\\ 
         &\textbf{DBMFIF \cite{zhang2024exploit} } &0.8703	&0.6959	&0.6620	&0.8169	&0.9048 &0.6642\\ 
         &\textbf{TCMOA \cite{zhu2024task} } &0.7545	&0.6034	&0.5307	&0.7874	&0.7681 &0.6475\\ 
         &\textbf{SwinMFF \cite{xie2024swinmff}  } &0.7259	&0.6312	&0.6336	&0.7445	&0.7285 &0.6050\\ 
         &\textbf{MCCSR-Net \cite{zheng2025CCSR} } &0.8891	&0.6925	&0.6898	&0.7903	&0.8961 &0.7233\\ 
         &\textbf{Ours} &\textbf{1.0887}&\textbf{0.7348} &\textbf{0.7642} &\textbf{0.8275} &0.9579 &0.7456\\
         \hline
         \multirow{9}{*}{\rotatebox{90}{\textbf{MFI-WHU \cite{zhang2021mff}}}} 
         &\textbf{SR \cite{yang2009multifocus}} & 1.1329 &0.7294 &0.7952 &0.8424 &0.9850 &0.8236 \\
         &\textbf{CSR \cite{liu2016image}} & 0.9802 &0.7157 &0.7897 &0.8423 &0.9427 &0.7827 \\
         &\textbf{MGIMF \cite{chen2021multi}} & 1.1269 &0.7151 &0.7782 &0.7973 &0.9825 &0.8157 \\
         &\textbf{QHOSVD \cite{miao2023quaternion}} &1.1507&0.7299	&0.8008	&0.8397	&0.9869	&0.8291\\
         &\textbf{SAMF \cite{li2024samf} }&\textbf{1.2155}&0.7310	&0.7980	&0.8366	&\textbf{0.9890}	&0.8268\\
         &\textbf{MFIF-GAN \cite{wang2021mfif}} &1.1648	&0.7343	&0.7994	&0.8410	&\textbf{0.9890}	&0.8304 \\ 
         &\textbf{ZMFF \cite{hu2023zmff}} &0.6933	&0.6342	&0.6705	&0.7915	&0.8540 &0.6860\\ 
         &\textbf{DBMFIF \cite{zhang2024exploit}} &1.0461	&0.7219	&0.7933	&0.8424	&0.9523 &0.7900\\ 
         &\textbf{TCMOA \cite{zhu2024task}} &0.8635&0.6389	&0.7026	&0.7748	&0.9139 &0.7655\\
         &\textbf{SwinMFF \cite{xie2024swinmff}  } &0.6921	&0.6776	&0.7436	&0.7906	&0.8406 &0.6599\\ 
         &\textbf{MCCSR-Net \cite{zheng2025CCSR}  } &1.2011	&0.7316	&0.7996	&0.8336	&0.9807 &0.8298\\ 
         &\textbf{Ours} &1.1743 &\textbf{0.7361} &\textbf{0.8017} &\textbf{0.8433} &0.9887 &\textbf{0.8318} \\
         \hline
    \end{tabular} }
    \label{tab:Lytro}
\end{table}

MFI-WHU dataset contains challenging small-area blurred structures. Our framework achieves the best fusion performance in $Q_{G}$ and $Q_{E}$ metrics, underscoring its effectiveness in detecting focus regions from small areas, particularly when compared to SAMF. Compared with QHOSVD and MCCSR-Net, QMCIF clearly preserves more accurate structural details and achieves the highest perceptual quality $Q_{CB}$.
However, in the $Q_{Y}$ metric, our framework performs slightly worse than GFDF and MFIF-GAN since both methods exhibit strong pixel-wise focus detection capabilities and our patch-wise fusion strategy tends to preserve less information from the original image in the boundary regions.

Extensive experiments on various datasets demonstrate that our framework significantly outperforms QHOSVD in terms of detail preservation and artifact removal under real-world challenging scenarios.

\textbf{Visual evaluation}.  To assess the effectiveness of focus detection and artifact suppression, we compare our framework against five recent state-of-the-art fusion approaches. More visual comparison results are seen in the supplementary material.

The top row of Fig. \ref{lytro05} presents the fusion results on the lytro dataset. As observed in the red bounding boxes, QHOSVD and TCMOA suffer from noticeable edge blurring in the building structures while MCCSR-Net introduces spatial artifacts in the same regions. In contrast, our framework produces a visually pleasing fusion result with clear object boundaries and no observable artifacts.

The middle row of Fig. \ref{lytro05} shows the results on the mffw dataset. The blue bounding boxes reveal that QHOSVD, SAMF, and TCMOA generate blurred pixels near the object boundaries, and the DBMFIF output contains pseudo-edges and ghosting artifacts in low-contrast regions. Our framework delivers the most visually satisfactory fusion result, effectively addressing the challenges posed by low-gradient backgrounds and uncertain boundary transitions.

The bottom row of Fig. \ref{lytro05} displays fusion results on the MFI-WHU dataset. This scene contains small and near-blurred structures such as electric wires and poles, making focus estimation particularly challenging. The red bounding boxes clearly show that QHOSVD, SAMF, TCMOA, and MCCSR-Net fail to preserve these fine structures. In contrast, our framework accurately preserves these focused regions, demonstrating its superior capability in fine-grained focus detection and high-fidelity structure retention.
Our framework can be easily extended to multiple color image fusion with architectural modifications. As an example, we extend our QMCIF framework to fuse three partially focused color images. By converting the input images into multiple quaternion representations, we obtain the detail layer and coefficient matrix for each input quaternion representation individually and generate patch-wise dual-scale focus maps using Eqs.~\eqref{eq-comparison-base} and~\eqref{eq-comparison-detail}. We select the patch with the most focus level of input images. Then we obtain dual-scale fusion results and the final fused result simply applying the QBDF and QSSR strategies. More details of multiple color image fusion are seen in the supplementary material.
\begin{figure}[htbp]
\centering
\includegraphics[width=1\columnwidth]{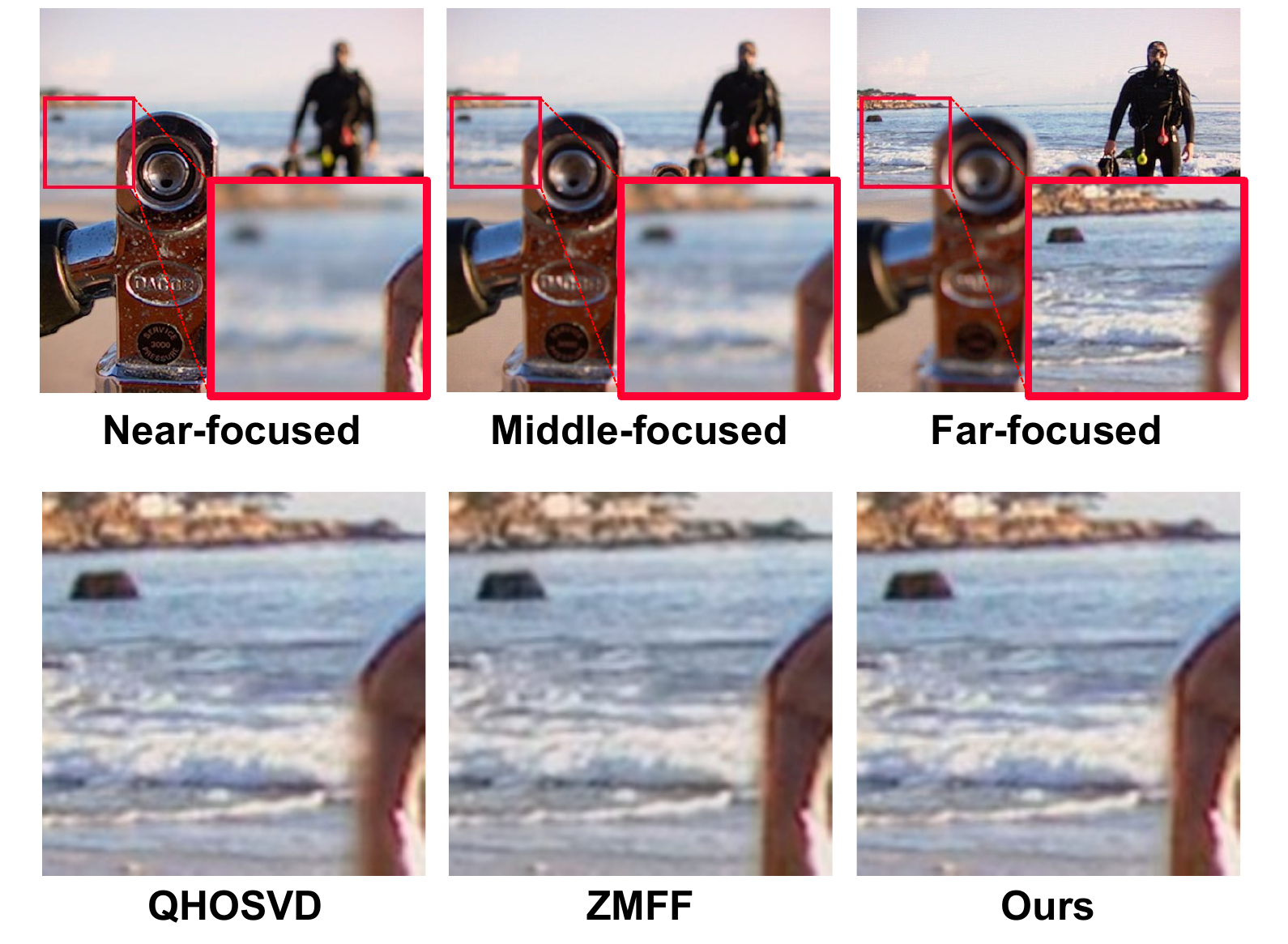}
\caption{Visual comparison of three partially focused color image fusion on lytro3 dataset.}
\label{lytro3}
\end{figure}
\begin{figure}[htbp]
\centering
\includegraphics[width=1\columnwidth]{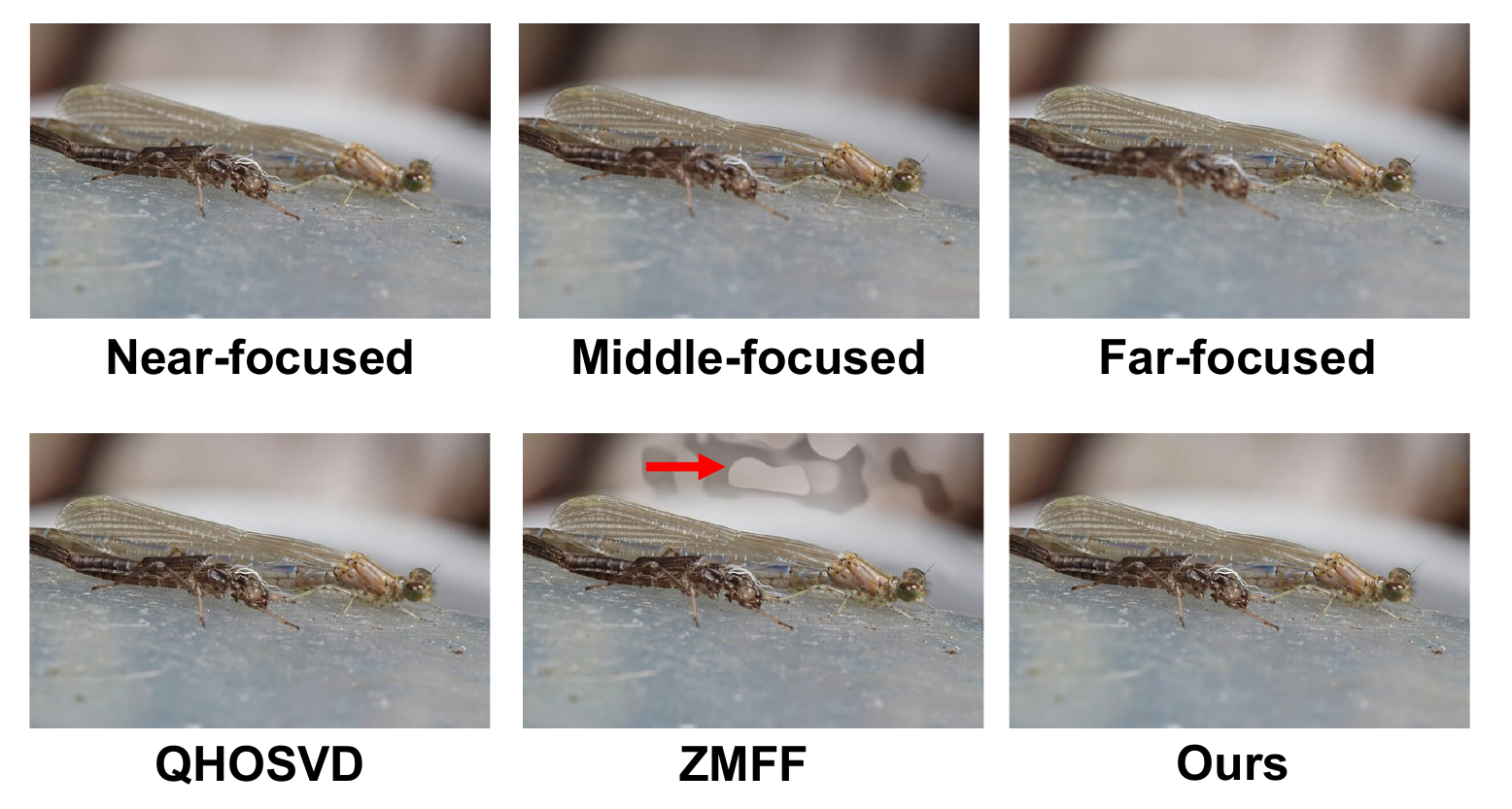}
\caption{Visual comparison of three partially focused color image fusion on mffw3 dataset.}
\label{mffw14}
\end{figure}

We conduct experiments on the lytro3 \cite{liu2015multi} and mffw3 \cite{xu2020mffw} datasets. We compare our framework with three representative approaches that are designed for multiple image fusion. 
The deep learning-based baseline ZMFF is used for this comparison \cite{hu2023zmff}. Figs. \ref{lytro3} and \ref{mffw14} presents the visual comparison results. More comparison results are included in the supplementary material. In Fig. \ref{lytro3}, QHOSVD produces visibly blurred shapes in the region marked by the red bounding box, especially around the device contour. ZMFF also suffers from color inconsistency in the same region, leading to degraded structure preservation and unnatural appearance. In Fig. \ref{mffw14}, ZMFF exhibits noticeable visual artifacts in the background, degrading perceptual quality. In contrast, our framework produces clean and structurally consistent results across all scenes. It effectively eliminates visual artifacts and maintains accurate focus representation in both foreground and background regions, as well as at structural boundaries. Our framework offers a simple yet robust solution for complex fusion tasks.

\subsection{Ablation study} \label{ablation}
In this subsection, we conduct an ablation study with different settings of our QMCIF framework on mffw dataset \cite{xu2020mffw}.
\begin{table}[hbtp]
\caption{Quantitative evaluation on different settings of our QMCIF framework. }
    \centering \resizebox{1\columnwidth}{!}{
    \begin{tabular}{|c|c|c|c|c|c|c|}
    \hline
          \multicolumn{2}{|c|}{Settings}&$Q_{MI} \uparrow$  &$Q_G \uparrow$  & $Q_P \uparrow$  & $Q_E \uparrow$ & $Q_{CB} \uparrow$ \\
         \hline
         \multirow{3}{*}{Patch size for QCAFD}&\textbf{$\boldsymbol{5\times5}$} &1.0503 &\textbf{0.7376} &0.7277 &0.8263 &0.7115 \\ 
         &\textbf{$\boldsymbol{10\times10}$} &1.0401 &0.7367 &0.7567 &\textbf{0.8276} &0.7276 \\
         &\textbf{$\boldsymbol{8\times8}$} &\textbf{1.0685} &0.7374 &\textbf{0.7832} &0.8234 &\textbf{0.7508} \\
         \hline
          \multirow{3}{*}{Fusion for QBDF and QSSR}&\textbf{QBDF-base-scale} &0.9312 &0.7021 &0.7113 &0.8099 &0.6830 \\
          &\textbf{QBDF-Detail-scale} &1.0388 &0.7357 &0.7457 &\textbf{0.8255} &0.7228 \\
         &\textbf{QSSR} &\textbf{1.0685} &\textbf{0.7374} &\textbf{0.7832} &0.8234 &\textbf{0.7508} \\ 
         \hline
         \multirow{2}{*}{Weight $\tau$ of QSSR}&\textbf{Same} &\textbf{1.0690} &0.7346 &0.7556 &0.8183 &0.7340 \\
         &\textbf{Adaptive} &1.0685 &\textbf{0.7374} &\textbf{0.7832} &\textbf{0.8234} &\textbf{0.7508} \\ 
         \hline
         \multirow{2}{*}{Representation of QMCIF}&\textbf{Real} &0.9555 &0.7082 &0.6849 &0.8043 &0.6748 \\ 
         &\textbf{Quaternion} &\textbf{1.0685} &\textbf{0.7374} &\textbf{0.7832} &\textbf{0.8234} &\textbf{0.7508} \\ 
         \hline
    \end{tabular} 
    }
    \label{tab:table100}
\end{table}

\textbf{Impact of patch size in QCAFD.} Table \ref{tab:table100} shows the quantitative analysis of different patch sizes within the QCAFD module. For QCAFD, we try different patch sizes to perform the QFED process and the patch-wise base-scale focus map generation for QBDF and QSSR strategies.
 The optimal performance is achieved with a patch size of $8 \times 8$, demonstrating balanced sensitivity to both high-texture and low-gradient regions.

\textbf{Effectiveness of QBDF and QSSR.} To verify the advantages of the QBDF and QSSR strategies, we perform the ablation experiments on the mffw data. This includes the first 10 pairs of partially focused color images under various exposures and obvious misalignment due to severe defocus spread effects. First, the coefficient matrix and the detail layer of quaternion representation of each input color image are extracted using Algorithm \ref{algorithm1}. Then, base-scale and detail-scale fused results are obtained using the dual-scale focus measures. These results are subsequently combined to generate the final output using the QSSR strategy. 
\begin{figure}[htbp]
\centering
\includegraphics[width=\columnwidth]{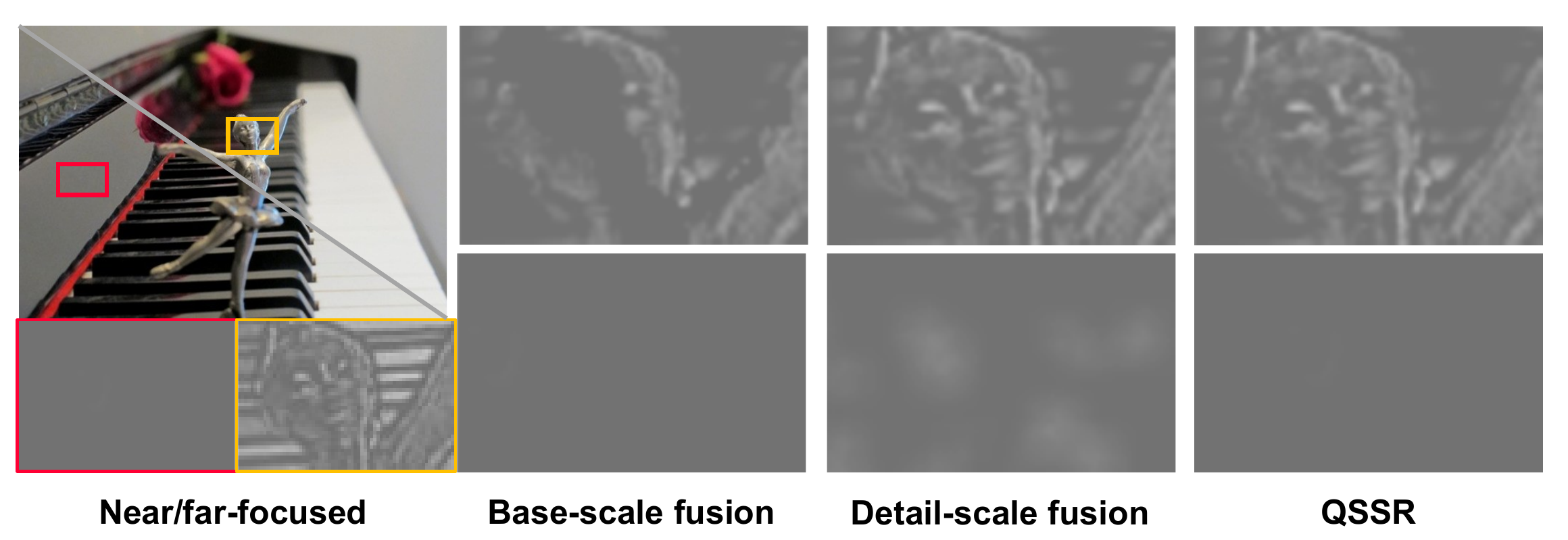}
\caption{Visual comparison of the background-focused input and various fused results including base-scale fusion, detail-scale fusion and the QSSR strategy. The first row represents the input and fused results. The second row represents the differences between the fused results and corresponding background-focused input.}
\label{fig:Adaptionmeasure}
\end{figure}
The importance of the QBDF and QSSR strategies is highlighted through qualitative and quantitative comparisons. Fig. \ref{fig:Adaptionmeasure} visually demonstrates while detail-scale fusion alone leads to artifacts and base-scale fusion smooths out fine details excessively. The QSSR strategy effectively preserves important structural details and eliminates artifacts, especially in boundary and low-gradient regions. The quantitative results in Table \ref{tab:table100} are align with this visual assessment. This indicates the improvements across multiple metrics when employing QSSR.

\textbf{Adaptive weighting in QSSR.} We further evaluate the robustness of adaptive weighting in the QSSR strategy (Eq. \eqref{adaptiveweights}). Table \ref{tab:table100} reveals that adaptive weighting consistently outperforms fixed weighting across all evaluated metrics. This confirms that the adaptive weighting accurately assesses local structural similarity to optimize patch selection.

\textbf{Effectiveness of QMCIF}. Table \ref{tab:table100} compares the quaternion representation with an equivalent real-domain under our framework. We replicate the fusion process in the real domain by converting color images to grayscale.
Quantitative results in Table \ref{tab:table100} highlight inherent advantages of quaternion representation in preserving inter-channel correlations and structural coherence during fusion.

\section{Conclusion} \label{conclusion}
This paper proposed a quaternion multi-focus color image fusion framework to perform high-quality color image fusion completely in the quaternion domain. This framework is flexible to be extended from fusing two color images at a time into fusing multiple color images simultaneously under complex challenging scenarios. This framework proposed a quaternion consistency-aware focus detection method. It consists of a quaternion focal element decomposition (QFED) module that jointly learns a low-rank structured coefficient matrix and a detail-scale layer and a dual-scale focus map generation strategy for robust focus detection. To solve the optimization problem of the QFED model, we developed an iterative algorithm under the framework of the quaternion alternating direction methods of multipliers. A quaternion base-detail fusion strategy was also introduced for high-quality base-scale and detail-scale color image fusion individually. A quaternion structural similarity strategy was further introduced to balance
the trade-off between focus information preservation and artifact removal. Our framework fuses two-scale quaternion matrices and compares them with the input quaternion representations to adaptively select the focused patches with a higher $WQ_{SSIM}$ value to form the final fused results. Extensive experiment
results have shown that our framework outperforms the state-of-the-art methods in terms of focus information preservation and artifact removal under real-world challenging scenarios.

\bibliographystyle{IEEEtran}
\bibliography{base}











\newpage

\vfill

\end{document}